\RequirePackage{fix-cm}
\documentclass[twocolumn]{svjour3}          
\smartqed  
\usepackage{graphicx}
\usepackage{latexsym}
\usepackage{url}
\usepackage{xcolor}
\definecolor{newcolor}{rgb}{.8,.349,.1}
\usepackage{float}
	\definecolor{blue(munsell)}{rgb}{0.0, 0.5, 0.69}
\usepackage{caption}
\usepackage{multirow}%
\usepackage{booktabs}
\usepackage{amssymb}
\usepackage[font={color=blue(munsell),bf},figurename=Fig.,labelfont={it}]{caption}
\usepackage{enumerate}
\usepackage{enumitem}
\usepackage{appendix}
\usepackage{algorithm}
\usepackage{algorithmicx}
\usepackage{algpseudocode}
\usepackage{hyperref}
\usepackage{amsmath}
\usepackage{colortbl}
\hypersetup{
     colorlinks = true,
     linkcolor = blue,
     anchorcolor = blue,
     citecolor = blue,
     filecolor = blue,
     urlcolor = blue
     }

\usepackage{array}
\usepackage{float}
\newcolumntype{A}{>{\centering\arraybackslash}m{1.2 cm}}
\newcolumntype{C}{>{\centering\arraybackslash}m{1.6 cm}}
\newcolumntype{E}{>{\centering\arraybackslash}m{1.4 cm}}
\begin{document}

\title{MiVID: Multi-Strategic Self-Supervision for Video Frame Interpolation using Diffusion Model    
}


\author{Priyansh Srivastava \and Romit Chatterjee \and Abir Sen \and Aradhana Behura \and Ratnakar Dash 
}
\institute{
Priyansh Srivastava \at
School of Computer Engineering, KIIT Deemed to be University, Bhubaneswar, Odisha, India
 \\
\email{priyansh0305@gmail.com}
\and
Romit Chatterjee \at
School of Computer Engineering, KIIT Deemed to be University, Bhubaneswar, Odisha, India\\
\email{chatterjeeromit86@gmail.com}
\and
\textbf{Abir Sen (Corresponding Author)} \at
School of Computer Engineering, KIIT Deemed to be University, Bhubaneswar, Odisha, India \\
\email{abir.senfcs@kiit.ac.in}
\and
Aradhana Behura \at
School of Computer Engineering, KIIT Deemed to be University, Bhubaneswar, Odisha, India \\
\email{aradhana.behurafcs@kiit.ac.in}
\and
Ratnakar Dash \at
Department of Computer Science and Engineering, National Institute of Technology, Rourkela, Odisha, 769008, India \\
\email{ratnakar@nitrkl.ac.in}
}

\date{Received: date / Accepted: date}
\maketitle

\abstract{
Video Frame Interpolation (VFI) remains a cornerstone in video enhancement, enabling temporal upscaling for tasks like slow-motion rendering, frame rate conversion, and video restoration. While classical methods rely on optical flow and learning-based models assume access to dense ground-truth, both struggle with occlusions, domain shifts, and ambiguous motion. This article introduces MiVID, a lightweight, self-supervised, diffusion-based framework for video interpolation. Our model eliminates the need for explicit motion estimation by combining a 3D U-Net backbone with transformer-style temporal attention, trained under a hybrid masking regime that simulates occlusions and motion uncertainty. The use of cosine-based progressive masking and adaptive loss scheduling allows our network to learn robust spatiotemporal representations without any high-frame-rate supervision.Our framework is evaluated on UCF101-7 and DAVIS-7 datasets. MiVID is trained entirely on CPU using the datasets and 9-frame video segments, making it a low-resource yet highly effective pipeline. Despite these constraints, our model achieves optimal results at just 50 epochs , competitive with several supervised baselines.This work demonstrates the power of self-supervised diffusion priors for temporally coherent frame synthesis and provides a scalable path toward accessible and generalizable VFI systems.
}

\keywords{Video Frame Interpolation, Diffusion Models, Temporal Attention, Frame
Masking, Self-Supervised Learning.}



\maketitle

\section{Introduction}\label{sec1}

%
Video Frame Interpolation (VFI) is an important task in computer vision. It creates intermediate frames between consecutive video frames, which improves temporal resolution. This has several applications, including slow-motion rendering, frame-rate upconversion \cite{lu2025self}, and video restoration \cite{shi2022transformer}. VFI is crucial in media production and streaming optimization. 

Traditional methods relying on optical flow, warping, and mixing \cite{xu2025inter} perform well in controlled settings, but struggle with occlusions, large movements, and complex lighting \cite{geetharamani2022hybrid}. Supervised learning methods that depend on accurate intermediate frames face significant challenges. They require large labeled datasets and often do not handle motion ambiguity well. In addition, they can produce artifacts or oversmooth the video, which limits their general use and scalability \cite{kye2025acevficomprehensivesurveyadvances}.

Recent deep learning techniques using convolutional or transformer-based models \cite{shi2022transformer} offer improvements in visual quality. However, they are still deterministic and require many resources, depending on high-frame-rate data that are densely sampled \cite{kim2025enhancing}. This limits their ability to adapt to different motions, leading to artifacts and lower generalization \cite{jain2024vidim}. The challenges of VFI have led to the exploration of generative approaches such as diffusion models. These models probabilistically capture motion changes through iterative denoising \cite{jain2024vidim}. However, current diffusion-based methods often require heavy sampling and complex multi-stage designs \cite{zheng2025eventdiff}.

To tackle these problems, this paper presents MiVID, a lightweight, self-supervised, diffusion-based framework for video frame interpolation. MiVID uses a compact 3D U-Net with temporal attention to capture spatiotemporal dynamics. It also employs a unified objective that combines several complementary losses to create stable and realistic frames. A progressive mask injection method gradually adds occlusions during training, which helps smooth the transition from rough reconstruction to fine perceptual details.

Unlike previous methods that rely on optical flow or require high computational power, MiVID uses diffusion principles and hybrid masking to provide accessible, scalable, and robust video interpolation, establishing a new benchmark for efficient generative video enhancement.
\par The remainder of this manuscript is structured as follows. In Section \ref{literature}, literature works related to diffusion models and self-supervised learning based frame interpolation. The proposed methodology is discussed in Section \ref{proposed}. Section \ref{experiment} deals with experimental details, dataset description, comparative analysis with other-state-of-the art works. Section \ref{discussion} describes the discussion portion. Finally, Section \ref{conclusion} includes the summary of the suggested work.

\section{Related Work}\label{literature}
Video Frame Interpolation (VFI) is a long-standing challenge in computer vision. It involves creating intermediate frames between video frames that are close in time. This topic connects motion estimation, temporal coherence, and generative modeling. Current research mostly fits into four groups: optical flow methods, deep learning interpolation networks, diffusion generative models, and self-supervised methods. Each of these has moved the field forward but faces issues with motion ambiguity, occlusion, and domain generalization. Our work seeks to connect these groups with a lightweight diffusion framework. This framework includes self-supervised temporal reasoning and occlusion-aware learning.

\vspace{-0.5cm}

\subsection{Optical Flow-Based Interpolation}

Classical methods estimate motion at the pixel level using optical flow. They then warp and blend frames to create intermediate results. Techniques like Super SloMo \cite{jiang2018super} and DAIN \cite{bao2019depth} improve motion estimation by using occlusion masks or depth cues. However, these methods struggle with occlusions, large motion, or complex lighting \cite{liu2020video}. They assume consistent brightness and flow, which does not hold true in real-world situations. This leads to ghosting or texture issues. Their reliance on high-frame-rate supervision and a single-ground-truth approach limits their scalability and flexibility \cite{sun2018pwcnetcnnsopticalflow}\cite{niklaus2018contextawaresynthesisvideoframe}\cite{ilg2016flownet20evolutionoptical}.

\subsection{Deep Learning-Based Interpolation Networks}

Deep learning models replace explicit flow with learned motion patterns. SepConv \cite{niklaus2017video} introduced kernel-based interpolation. AdaCoF \cite{lee2020adacof} added deformable convolutions for more flexible motion handling. 3D CNNs like FLAVR \cite{kalluri2020flavr} and geometry-aware models such as QVI \cite{xu2019quadratic} capture spatiotemporal dynamics more effectively. However, deterministic loss functions (L1/L2) lead to oversmoothing and weak temporal coherence. These models need dense, high-frame-rate data and do not perform well with real-world motion or changes in lighting \cite{shi2022transformer}. While they achieve strong results in tests, they struggle with motion ambiguity and occlusion.

\vspace{-0.5cm}

\subsection{Diffusion Models for Temporal Generation}
Diffusion models have recently become important for video generation by modeling complex multimodal distributions \cite{lyu2025tlbvfitemporalawarelatentbrownian}\cite{lew2024disentangledmotionmodelingvideo}. They gradually add and remove noise to learn generative transitions without explicit flow. VIDIM \cite{jain2024vidim} introduced cascaded diffusion for multi-frame synthesis, and LDMVFI \cite{li2024ldmvfi} cut computing costs using latent diffusion. However, these models often need multi-stage training and remain heavy on resources. Other variants, like MADiff \cite{li2023madiff} and EDEN \cite{zheng2024eden}, include motion conditioning or noise scheduling but still rely on dense supervision and experience temporal instability. Our work simplifies this by designing a single-stage, CPU-efficient diffusion system trained without dense supervision.

\vspace{-0.5cm}

\subsection{Self-Supervised learning based Frame Interpolation}

Self-supervised methods tackle the annotation problem by creating proxy goals based on temporal consistency or masking. Reda et al. \cite{reda2019self} introduced time-cycle consistency, while others focus on masked frame reconstruction to learn motion dynamics \cite{wei2023maskedfeaturepredictionselfsupervised}. These models eliminate the need for ground truth but remain predictable and sensitive to non-linear motion or lighting changes.

To tackle these challenges, we combine self-supervised learning with diffusion modeling using a hybrid masking strategy that blends temporal dropout and motion-aware occlusion. This approach allows for effective denoising and training on standard datasets like DAVIS-7 and UCF101-7 without needing intermediate labels. As a result, we achieve efficient, high-quality, and general video frame interpolation for real-world situations.

\vspace{-0.4cm}
\section{Proposed Methodology}\label{proposed}

\begin{figure*}[htbp!]
\centerline{\includegraphics[height=3cm, width =\textwidth]{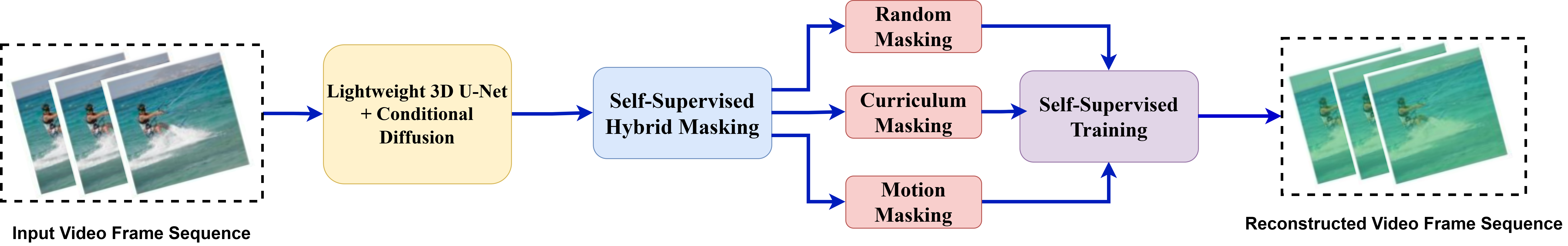}}
\centering
\vspace{0.5cm}
\caption{Architecture of our proposed MiVID workflow.}
\label{fig1_3}
\end{figure*}

This section introduces our self-supervised video frame interpolation framework, which integrates a conditional diffusion process~\cite{ho2020denoising}, temporal attention modules~\cite{shi2022transformer}, and multi-strategic masking~\cite{reda2019self}. Our approach is explicitly designed to operate without dense supervision or explicit optical flow, enabling it to generalize across diverse motion regimes and occlusion-heavy scenarios. The core of our method is a lightweight, spatiotemporal U-Net backbone~\cite{ronneberger2015u} augmented with attention along the temporal axis, allowing dynamic context modeling across video frames.

In contrast to cascaded or latent diffusion approaches that often require multi-stage training, handcrafted encoders, or super-resolution modules~\cite{jain2024vidim}\cite{li2024ldmvfi}, our single-stage architecture directly learns to denoise and reconstruct masked intermediate frames using raw RGB input. A hybrid masking mechanism—combining random temporal masking and motion-aware occlusion simulation~\cite{reda2019self} enables self-supervised learning on conventional datasets. This combination of conditional generative modeling and structured masking forces the model to infer plausible motion trajectories while preserving high-frequency detail, enabling it to perform accurate, high-fidelity frame interpolation under real-world conditions.

Compared to existing diffusion-based VFI models, our proposed framework in Fig.~\ref{fig1_3} is notably resource-efficient and can be trained on CPU hardware, making it accessible for broader research and application. The integration of curriculum masking and progressive mask injection further enhances robustness, allowing the model to adaptively focus on both coarse structure and fine perceptual details as training progresses. The stages of our framework have been discussed as follows:
\vspace{-0.4cm}
\subsection{Model Overview: Mathematical Formulation}
Given a video sequence containing $T$ frames, our objective is to synthesize one or more temporally intermediate frames $\hat{I}_t$ between two known inputs. We treat this as a conditional masked reconstruction problem within a denoising diffusion framework~\cite{ho2020denoising}, where a subset of frames are masked and replaced with noise. The model is trained to denoise them using spatiotemporal context.

Our model architecture follows a U-Net-like encoder and decoder structure ~\cite{ronneberger2015u} equipped with temporal attention~\cite{shi2022transformer} and noise conditioning.


Let a video sequence be 
\[
x = \{I_1, I_2, \dots, I_T\}, \quad I_t \in \mathbb{R}^{C \times H \times W}.
\]

Define a masking operator $M \in \{0,1\}^T$ such that $M_t = 1$ if frame $t$ is masked (to be reconstructed), and $M_t = 0$ otherwise.  
The visible frames are $(1-M) \odot x$, while masked frames are $M \odot x$.
The pipeline consists of four key components:

\begin{itemize}[label=\textbullet]
    \item \textbf{Encoder:} A compact 3D convolutional encoder extracts low-level motion-aware features from the input video segment. It processes tensors of shape $[B, C, T, H, W]$ via stacked 3D convolutions with GELU activations, reducing dimensionality while preserving temporal coherence. These features capture localized spatiotemporal structures necessary for reconstruction.
    
\begin{equation*}
h = \mathcal{E}_\phi(z_t,\, (1-M)\odot x_0,\, t),
\tag{1}
\end{equation*}

    \item \textbf{Temporal Attention Block:} Positioned at the bottleneck, this transformer-style module applies scaled dot-product attention over temporal tokens~\cite{shi2022transformer} to facilitate long-range temporal reasoning. The attention weights are computed across frames, allowing the model to focus on relevant timepoints and motion paths, and implicitly learn occlusion relationships and motion trajectories.
    
\begin{equation*}
\mathrm{Attn}(Q,K,V) = \mathrm{softmax}\!\left(\frac{QK^\top}{\sqrt{d}}\right) V,
\tag{2}
\end{equation*}

\begin{equation*}
\tilde{h}^{(1:T)} = \mathrm{Attn}\big(W_Q h^{(1:T)}, W_K h^{(1:T)}, W_V h^{(1:T)}\big).
\tag{3}
\end{equation*}
    
    \item \textbf{Decoder:} A symmetric decoder composed of 3D convolutions progressively upsamples and refines the latent features. Skip connections are applied from encoder layers to preserve fine spatial details and stabilize training.
\begin{equation*}
\hat{x}_{\text{latent}} = \mathcal{D}_\psi(\tilde{h}, \{s_i\}_i).
\tag{4}
\end{equation*}
    
    The final output head predicts the residual noise $\hat{\epsilon}_\theta$ corresponding to a noisy intermediate frame latent $z_t$. The predicted noise is used in the reverse denoising process to reconstruct the clean latent $\hat{x}_0$.
\end{itemize}

\begin{itemize}[label=\textbullet]
    \item \textbf{Forward Diffusion (noising process):}
At timestep $t \in \{1,\dots,T_{\max}\}$, with variance schedule $\{\alpha_t\}$ and cumulative product $\bar{\alpha}_t = \prod_{s=1}^t \alpha_s$, we inject Gaussian noise $\varepsilon \sim \mathcal{N}(0,I)$ only into masked frames:
\begin{equation*}
z_t = (1-M) \odot x_0 \;+\; M \odot \Big( \sqrt{\bar{\alpha}_t} \, x_0 + \sqrt{1-\bar{\alpha}_t} \, \varepsilon \Big).
\tag{5}
\end{equation*}

\item \textbf{Noise Prediction and Reconstruction:}
\begin{equation*}
\hat{\varepsilon} = \varepsilon_\theta(z_t,\, t,\,(1-M)\odot x_0),
\tag{6}
\end{equation*}
\begin{equation*}
\hat{x}_0 = \frac{1}{\sqrt{\bar{\alpha}_t}}\Big(z_t - \sqrt{1-\bar{\alpha}_t}\,\hat{\varepsilon}\Big).
\tag{7}
\end{equation*}
\begin{equation*}
\hat{x} = (1-M)\odot x_0 + M \odot \hat{x}_0.
\tag{8}
\end{equation*}

\item \textbf{Training Objective:}
\begin{equation*}
\mathcal{L}(\theta,\phi,\psi) =
\mathbb{E}_{x_0,\,t,\,\varepsilon}\;
\Big\| \varepsilon - \varepsilon_\theta(z_t, t,(1-M)\odot x_0) \Big\|_2^2.
\tag{9}
\end{equation*}

\item \textbf{Reverse Process (Sampling):}
\begin{equation*}
\hat{\varepsilon}_t = \varepsilon_\theta(z_t, t,(1-M)\odot x_0),
\tag{10}
\end{equation*}
\begin{equation*}
\hat{x}_{0,t} = \frac{1}{\sqrt{\bar{\alpha}_t}}
\Big(z_t - \sqrt{1-\bar{\alpha}_t}\,\hat{\varepsilon}_t\Big),
\tag{11}
\end{equation*}
\begin{equation*}
\mu_{t-1} = 
\frac{\sqrt{\bar{\alpha}_{t-1}} \, \beta_t}{1-\bar{\alpha}_t}\,\hat{x}_{0,t}
+ \frac{\sqrt{\alpha_t}(1-\bar{\alpha}_{t-1})}{1-\bar{\alpha}_t}\,z_t,
\tag{12}
\end{equation*}
\begin{equation*}
z_{t-1} = \mu_{t-1} + \sigma_t \zeta, \quad \zeta \sim \mathcal{N}(0,I),
\tag{13}
\end{equation*}
\begin{equation*}
z_{t-1} \leftarrow (1-M)\odot x_0 + M \odot z_{t-1}.
\tag{14}
\end{equation*}
\end{itemize}

In the forward diffusion process, where clean frames $x_0$ are progressively noised.  The $M$ masks the unknown frames, ensuring only missing regions are corrupted.  The $\bar{\alpha}_t = \prod_{i=1}^t \alpha_i$ (with $\alpha_i = 1-\beta_i$) controls the noise schedule.   This produces training data where the network learns to denoise masked video frames.


The reverse denoising phase reconstructs the clean video estimate $\hat{x}_{0,t}$ from the noisy latent $z_t$.  The $\hat{\varepsilon}_t$ is the predicted noise from the model.  It acts as the inverse of the forward diffusion (Eq.~1) and provides a denoised reference used in the reverse step.





This objective $\mathcal{L}$ trains the model to predict the Gaussian noise $\varepsilon$ that is introduced during the forward process. 
It is formulated as a simple mean squared error (MSE) loss, 
which enforces accurate noise estimation. Once optimized, the model can apply the reverse process to reconstruct clean intermediate frames.


During training, the network receives a partially masked video segment (via random and motion-guided masking strategies~\cite{reda2019self}). A forward diffusion process injects Gaussian noise into the masked frames to simulate degradation~\cite{ho2020denoising}. The model learns to conditionally recover these frames by denoising the latent representation, with guidance from the visible frames.

\subsection{Conditional Diffusion Formulation}

The proposed method is based on the Denoising Diffusion Probabilistic Model (DDPM)~\cite{ho2020denoising}, which defines a two-stage process: a forward process that gradually corrupts the ground truth frame with Gaussian noise, and a learned reverse process that reconstructs the clean frame by iteratively denoising.

Let $x_0$ represent the clean latent representation of a masked intermediate frame, and let $z_t$ denote its noisy version at timestep $t$. The forward diffusion process is defined as:

\begin{equation*}
q(z_t \mid x_0) = \mathcal{N}\left(z_t; \sqrt{\bar{\alpha}_t} x_0, (1 - \bar{\alpha}_t)\mathbf{I} \right)
\tag{15}
\end{equation*}

where $\bar{\alpha}_t = \prod_{s=1}^{t} \alpha_s$, and $\alpha_t = 1 - \beta_t$ denotes the noise schedule~\cite{ho2020denoising}. This formulation ensures that as $t \rightarrow T$, the sample $z_t$ becomes increasingly close to Gaussian noise.

To reverse this process, a neural network $\epsilon_\theta(z_t, t, c)$ is trained to predict the noise component $\epsilon$ added at each timestep $t$, conditioned on features $c$ extracted from the visible (unmasked) context frames:

\begin{equation*}
\mathcal{L}_{\text{diff}} = \mathbb{E}_{x_0, \epsilon, t} \left[ \left\| \epsilon - \epsilon_\theta(z_t, t, c) \right\|_2^2 \right].
\tag{16}
\end{equation*}

The conditioning features $c$ are derived from the encoder and temporal attention modules applied to the visible frames~\cite{shi2022transformer}. During inference, we sample $z_T \sim \mathcal{N}(0, \mathbf{I})$ and apply the reverse process iteratively to obtain the denoised latent $x_0$:

\begin{equation*}
z_{t-1} = \frac{1}{\sqrt{\alpha_t}} \left( z_t - \frac{1 - \alpha_t}{\sqrt{1 - \bar{\alpha}_t}} \cdot \epsilon_\theta(z_t, t, c) \right)
\tag{17}
\end{equation*}

The clean latent $x_0$ is then decoded into the final interpolated frame $\hat{I}_t$.

This conditional denoising strategy allows our model to synthesize realistic and motion-consistent intermediate frames even in the presence of occlusions, abrupt motion, and noisy input contexts. The design also enables flexible inference, as the number and position of interpolated frames can be adjusted dynamically at test time.

\subsection{Temporal Attention Mechanism}

We insert temporal attention layers at the bottleneck to explicitly model long-range motion dependencies~\cite{shi2022transformer}. Given spatiotemporal features $f \in \mathbb{R}^{B \times T \times C \times H \times W}$, we flatten the spatial dimensions and compute attention weights:

\begin{equation*}
\text{Attention}(Q, K, V) = \text{Softmax} \left( \frac{QK^\top}{\sqrt{d_k}} \right) V
\tag{18}
\end{equation*}
where $Q$, $K$, and $V$ are linear projections of the input along the temporal dimension. This helps the model infer motion-aware correspondences across time without explicitly computing optical flow. By leveraging self-attention, the model can capture both short-term and long-term dependencies, which is crucial for handling complex motion and occlusions in real-world videos.

\subsection{Self-Supervised Masking Strategy}

To train without ground-truth intermediate frames, we adopt a multi-strategic masking scheme~\cite{reda2019self}.



Let $x = \{I_1, I_2, \dots, I_T\}$ denote a video sequence of $T$ frames.  
This model define a masking operator $M \in \{0,1\}^{T}$, where $M_t = 1$ indicates that frame $I_t$ is masked and replaced with noise or a constant value.  

\subsubsection{Random Masking}
 Randomly selected intermediate frames are replaced by noise or a constant value to simulate occlusion.
Randomly select a subset of intermediate frames to mask:
\begin{equation*}
M_t \sim \text{Bernoulli} (p_r), \quad t \in \{2, \dots, T-1\}
\tag{19}
\end{equation*}
\begin{equation*}
\resizebox{0.8\columnwidth}{!}{$
\tilde{x}_t = (1-M_t) I_t + M_t \eta_t, \quad \eta_t \sim \mathcal{N}(0,I)
$}
\tag{20}
\end{equation*}
 Each frame is masked independently with probability $p_r$, simulating random occlusion.

\subsubsection{Motion-Guided Masking}
Motion-dense regions are detected via inter-frame differences, and masking is biased toward those frames.
Compute motion intensity between consecutive frames:
\begin{equation*}
\Delta_t = \| I_t - I_{t-1} \|_2, \quad t = 2, \dots, T
\tag{21}
\end{equation*}

Mask frames proportionally to motion magnitude:
\begin{equation*}
M_t \sim \text{Bernoulli}\Big(p_m \frac{\Delta_t}{\sum_{k=2}^{T} \Delta_k} \Big)
\tag{22}
\end{equation*}
\begin{equation*}
\tilde{x}_t = (1-M_t) I_t + M_t \eta_t
\tag{23}
\end{equation*}

Frames with higher motion are more likely to be masked, encouraging the model to learn motion completion.

\subsubsection{Curriculum Masking}
Gradually increase the overall masking ratio during training:
\begin{equation*}
p_{\text{mask}}(e) = p_{\min} + (p_{\max}-p_{\min}) \frac{e}{E_{\max}}
\tag{24}
\end{equation*}
where $e$ is the current training epoch and $E_{\max}$ is the total number of epochs. Then sample mask:
\begin{equation*}
M_t \sim \text{Bernoulli}(p_{\text{mask}}(e)), \quad t \in \{2, \dots, T-1\}
\tag{25}
\end{equation*}

 This curriculum allows the model to first learn simple reconstructions with fewer masked frames, and progressively handle more challenging occlusions.

\subsubsection{Hybrid Masking}

Combine all three strategies into a single mask:
\begin{equation*}
M_t = M_t^{\text{random}} \; \vee \; M_t^{\text{motion}} \; \vee \; M_t^{\text{curriculum}}
\tag{26}
\end{equation*}
\begin{equation*}
\tilde{x}_t = (1-M_t) I_t + M_t \eta_t
\tag{27}
\end{equation*}

This final hybrid approach enforces temporal reasoning and motion completion without requiring ground-truth high-frame-rate intermediate frames. Masking probabilities can be adapted to dataset characteristics or motion statistics.


\subsection{Custom Training Objective}

Our training objective is composed of multiple complementary losses that jointly supervise the generation process. These include a diffusion loss for guiding the denoising process and several image-based reconstruction losses to ensure fidelity, perceptual similarity, and realism. The overall training loss is formulated as:

\begin{equation*}
\mathcal{L}_{\text{total}} = \mathcal{L}_{\text{diff}} + \mathcal{L}_{\text{pix}} + \mathcal{L}_{\text{perc}} + \mathcal{L}_{\text{lpips}}
\tag{28}
\end{equation*} 
Each component is detailed below:

\paragraph{\textbf{Diffusion Loss $\mathcal{L}_{\text{diff}}$:}}  
This loss supervises the core denoising network of the diffusion model. Following the standard DDPM training protocol~\cite{ho2020denoising}, we predict the noise $\epsilon$ added to a clean input $x_0$ at a random timestep $t$, and minimize the mean squared error between the ground-truth noise and the predicted noise $\epsilon_\theta$. Formally,
\begin{equation*}
\mathcal{L}_{\text{diff}} = \mathbb{E}_{x_0, \epsilon, t} \left[ \left\| \epsilon - \epsilon_\theta(z_t, t, f) \right\|_2^2 \right]
\tag{29}
\end{equation*}
where $z_t$ is the noisy latent at timestep $t$ and $f$ denotes any conditioning features such as adjacent frames.

\paragraph{\textbf{Pixel Reconstruction Loss $\mathcal{L}_{\text{pix}}$:}}  
To encourage accurate reconstruction of the target frame at the pixel level, we combine both mean squared error (MSE) and L1 loss\cite{liu2018futureframepredictionanomaly}. MSE penalizes large deviations, while L1 is more robust to outliers and helps preserve edges:
\begin{equation*}
\mathcal{L}_{\text{pix}} = \lambda_{\text{mse}} \cdot \| \hat{x}_0 - x_{\text{target}} \|_2^2 + \lambda_{\text{l1}} \cdot \| \hat{x}_0 - x_{\text{target}} \|_1
\tag{30}
\end{equation*}
where $\hat{x}_0$ is the predicted clean frame, and $x_{\text{target}}$ is the ground truth.

\paragraph{\textbf{Perceptual Loss $\mathcal{L}_{\text{perc}}$:}}  
Low-level pixel differences often fail to capture high-level semantic similarity. To address this, we incorporate a perceptual loss based on the VGG network~\cite{zhang2018unreasonable}, which compares feature activations:
\begin{equation*}
\mathcal{L}_{\text{perc}} = \lambda_{\text{perc}} \cdot \| \phi(\hat{x}_0) - \phi(x_{\text{target}}) \|_1
\tag{31}
\end{equation*}
where $\phi(\cdot)$ denotes activations from a pre-trained VGG network, typically taken from intermediate layers.

\paragraph{\textbf{LPIPS Loss $\mathcal{L}_{\text{lpips}}$:}}  
To further enhance perceptual quality, we employ the Learned Perceptual Image Patch Similarity (LPIPS) loss~\cite{zhang2018unreasonable}, which better correlates with human visual perception compared to traditional metrics:
\begin{equation*}
\mathcal{L}_{\text{lpips}} = \lambda_{\text{lpips}} \cdot \text{LPIPS}(\hat{x}_0, x_{\text{target}})
\tag{32}
\end{equation*}

\subsection{Training and Inference Workflow}

The overall pipeline for our method is summarized in Algorithm~\ref{alg:proposed_vfi}. During training, the model receives partially masked video segments, applies noise through the forward diffusion process, and learns to reconstruct the masked frames by leveraging visible context and temporal attention. The loss function combines generative, pixel-wise, and perceptual objectives to ensure both fidelity and realism. For inference, the model synthesizes intermediate frames by iteratively denoising random noise, conditioned on the available input frames and learned spatiotemporal priors. This unified procedure enables robust frame interpolation on diverse video content without requiring dense supervision or optical flow.

\begin{algorithm}
\caption{Self-Supervised Video Frame Interpolation via Conditional Diffusion}
\label{alg:proposed_vfi}
\begin{algorithmic}[1]
\Require Video segment $x = \{I_1, I_2, \dots, I_T\}$, Diffusion steps $T_d$, Loss weights $\lambda$,Combined Mask $\mathcal{M}$
\Ensure Reconstructed masked frames $\hat{I}_t$

\Statex \Comment{\textbf{Training Phase}}
\For{each training epoch}
    \For{each video batch $x$}
        \State Mask frames: $(x_{\text{masked}}, x_{\text{target}}, \mathcal{M})$
        \State Sample timestep $t \sim \mathcal{U}(1, T_d)$
        \State Add noise: $z_t = \sqrt{\bar{\alpha}_t} x_{\text{target}} + \sqrt{1 - \bar{\alpha}_t} \cdot \epsilon$
        \State Encode: $f = \texttt{Encoder}(x_{\text{masked}})$
        \State Apply attention: $f' = \texttt{TemporalAttention}(f)$
        \State Predict noise: $\hat{\epsilon}_\theta = \epsilon_\theta(z_t, t, f')$
        \State Decode: $\hat{x}_0 = \texttt{Decoder}(z_0)$
        \State Compute total loss $\mathcal{L}_{\text{total}}$ (Eq. 7)
        \State Backpropagate and update $\theta$
    \EndFor
\EndFor

\Statex \Comment{\textbf{Inference Phase}}
\State Given $I_0$, $I_1$: extract $f \gets \texttt{Encoder}([I_0, I_1])$
\State $f' \gets \texttt{TemporalAttention}(f)$
\State Sample $z_T \sim \mathcal{N}(0, \mathbf{I})$
\For{$t = T_d, \dots, 1$}
    \State Predict noise: $\hat{\epsilon}_\theta = \epsilon_\theta(z_t, t, f')$
    \State Denoise: 
    \[
    z_{t-1} = \frac{1}{\sqrt{\alpha_t}} \left( z_t - \frac{1 - \alpha_t}{\sqrt{1 - \bar{\alpha}_t}} \cdot \hat{\epsilon}_\theta \right)
    \]
\EndFor
\State $\hat{I}_{0.5} \gets \texttt{Decoder}(z_0)$
\end{algorithmic}
\end{algorithm}

\vspace{2mm}
\noindent

\section{Experimental Details}\label{experiment}

To evaluate the performance and generalization of the proposed MiVID framework, we conducted experiments on two difficult video interpolation benchmarks: UCF101-7 \cite{soomro2012ucf101} and DAVIS-7 \cite{davis2017}.

\subsection{Datasets description}

\textbf{UCF101-7} comes from the UCF101 action recognition dataset. This dataset has over 13,000 YouTube clips across 101 categories of human actions. UCF101-7 samples seven consecutive frames from chosen videos, resulting in about 400 short clips. These clips capture various motion dynamics while keeping the original 320×240 resolution and a frame rate of 25 fps. This dataset focuses on complex human movements and serves as a compact but tough benchmark for testing temporal modeling and motion synthesis.

\textbf{DAVIS-7} is adapted from the DAVIS video object segmentation dataset. This dataset provides high-quality sequences with rich motion, occlusion, and complex segmentation. Each sequence consists of seven consecutive frames pulled from the original DAVIS videos, maintaining realistic object interactions and scene changes. With resolutions reaching up to 4K and frame rates around 24 fps, DAVIS-7 challenges a model’s ability to preserve spatial structure and temporal coherence amid difficult motion and appearance changes.

\subsection{Experimental Setup}
Experiments were run on Google Colab Pro with v2-8 TPU and 32GB RAM, and select runs on an Intel Xeon CPU (32 cores, 128GB RAM). We use PyTorch 2.2 and CUDA 12. Training uses the Adam optimizer ($\beta_1=0.9, \beta_2=0.999$), initial learning rate $10^{-4}$ with cosine annealing, and batch size 8 for 100 epochs. Random seed is fixed to 42 for reproducibility. \\

\begin{figure}[h!]
    \raggedright
    \includegraphics[width=0.95\columnwidth]{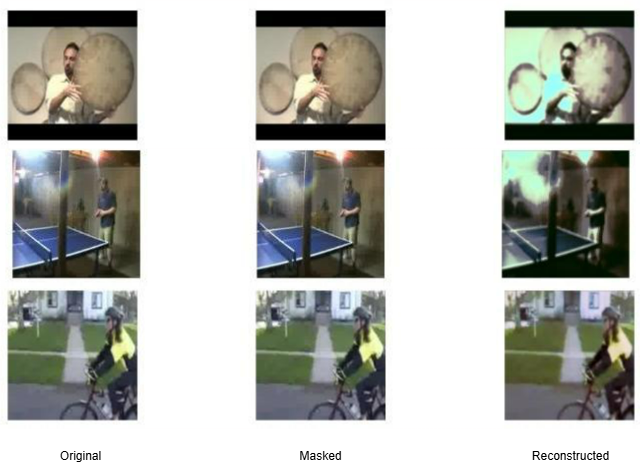}
    \captionsetup{justification=raggedright, singlelinecheck=false}
    \caption{Reconstruction output of a sample from the UCF101-7 dataset. The order of images in the figure denotes Original, Masked, and Reconstructed frames respectively.}
\label{ucf_recon}
\end{figure}

\begin{figure}[h!]
    \raggedright
    \includegraphics[width=0.95\columnwidth]{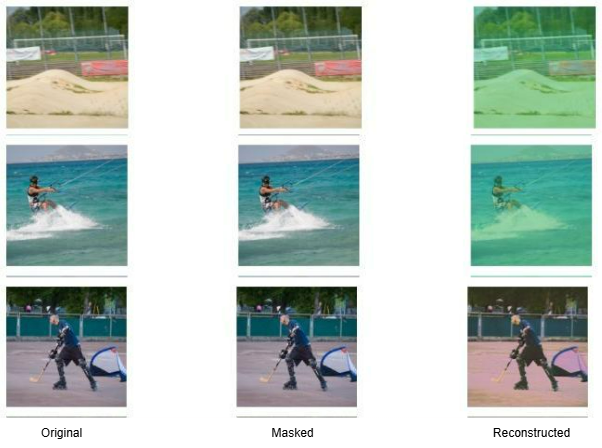}
    \captionsetup{justification=raggedright, singlelinecheck=false}
    \caption{Reconstruction output of a sample from the DAVIS-7 dataset. The order of images in the figure denotes Original, Masked, and Reconstructed frames respectively.}
    \label{davis_recon}
\end{figure}

\begin{figure}[h!]
    \centering
        \includegraphics[width=9 cm, height=8 cm]{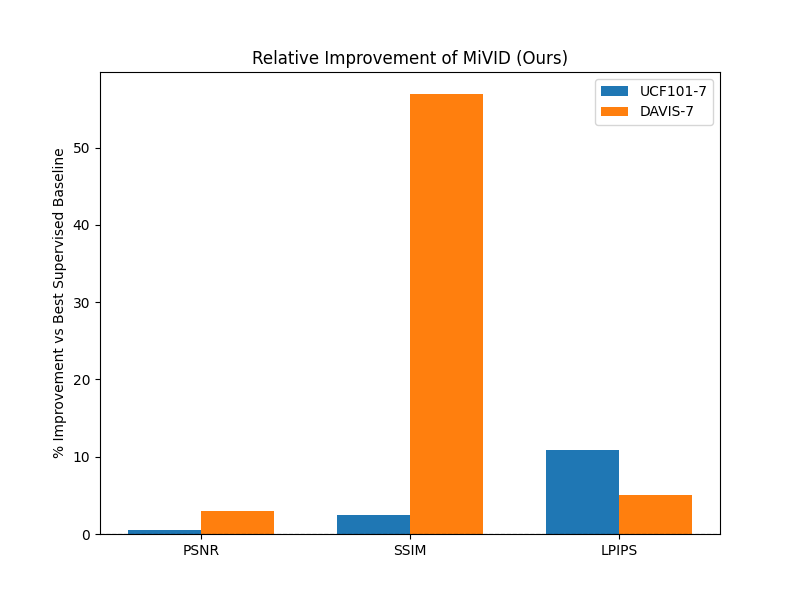}
    \caption{The figure shows the relative improvements of MiVID over the best supervised baselines on UCF101-7 and DAVIS-7 datasets. On UCF101-7, MiVID achieves an advantage of 0.7$\%$ in PSNR, 2.6$\%$ in SSIM, and 10.9$\%$ in LPIPS. On DAVIS-7, the improvements are more pronounced, with 3.0$\%$ in PSNR, 57.0$\%$ in SSIM, and 5.0$\%$ in LPIPS.}
\label{comparison}
\end{figure}
\vspace{-0.5cm}
\subsection{Evaluation Metrics}

To comprehensively assess the quality of interpolated frames, we employ three complementary metrics that capture different aspects of image quality: pixel-level fidelity, structural similarity, and perceptual realism.

\textbf{Peak Signal-to-Noise Ratio (PSNR)} measures pixel-level reconstruction accuracy by comparing the mean squared error between predicted and ground truth frames:

\begin{equation*}
\text{PSNR} = 10 \log_{10}\left(\frac{L^2}{\text{MSE}}\right)
\tag{33}
\end{equation*}

where $L$ is the maximum possible pixel value (255 for 8-bit images), MSE denotes the mean square error.

\begin{equation*}
\text{MSE} = \frac{1}{HW} \sum_{i=1}^{H} \sum_{j=1}^{W} (I_{\text{pred}}(i,j) - I_{\text{gt}}(i,j))^2
\tag{34}
\end{equation*}

Higher PSNR values indicate better reconstruction fidelity, with typical ranges of 20-40 dB for video interpolation tasks.\\

\textbf{Structural Similarity Index Measure (SSIM)} evaluates structural information preservation by comparing luminance, contrast, and structure:

\begin{equation*}
\text{SSIM}(x,y) = \frac{(2\mu_x\mu_y + c_1)(2\sigma_{xy} + c_2)}{(\mu_x^2 + \mu_y^2 + c_1)(\sigma_x^2 + \sigma_y^2 + c_2)}
\tag{35}
\end{equation*}

where $\mu_x, \mu_y$ are mean intensities, $\sigma_x^2, \sigma_y^2$ are variances, $\sigma_{xy}$ is covariance, and $c_1, c_2$ are stability constants. SSIM ranges from 0 to 1, with values closer to 1 indicating better structural similarity.\\

\textbf{Learned Perceptual Image Patch Similarity (LPIPS)} measures perceptual distance using deep features from pre-trained networks:

\begin{equation*}
\text{LPIPS} = \sum_{l} \frac{1}{H_l W_l} \sum_{h,w} \|w_l \odot (\hat{y}_{hw}^l - y_{hw}^l)\|_2^2
\tag{36}
\end{equation*}

where $\hat{y}^l, y^l$ are normalized activations from network layer $l$, and $w_l$ are learned weights. Lower LPIPS values indicate better perceptual quality, with typical ranges of 0.1-0.3 for high-quality interpolation.

These metrics collectively provide a comprehensive evaluation framework: PSNR captures pixel-accurate reconstruction, SSIM measures structural preservation, and LPIPS evaluates perceptual similarity aligned with human visual perception. All metrics are computed on the center interpolated frames and averaged across the entire test dataset.
\vspace{-0.4cm}
\subsection{Experimental Results}

We evaluate the proposed MiVID framework on two widely adopted benchmarks, UCF101-7 and DAVIS-7, and compare against state-of-the-art supervised video frame interpolation (VFI) methods including RIFE~\cite{huang2022rife}, FILM~\cite{reda2022film}, LDMVFI~\cite{danier2023ldmvfi},VIDIM~\cite{jain2024vidim} and AMT~\cite{jiang2018superslomo}. \\

\begin{table}[ht]
    \captionsetup{justification=raggedright,singlelinecheck=false}
    \caption{Performance comparison of our model with baselines on UCF101-7.}
    \label{tab:results1}
    \resizebox{\columnwidth}{!}{
    \begin{tabular}{lcccccc}
        \toprule
        \textbf{Model} & \textbf{Supervision} & \textbf{Dataset} & \textbf{PSNR↑} & \textbf{SSIM↑} & \textbf{LPIPS↓} \\
        \midrule
        AMT~\cite{jiang2018superslomo} & Supervised & UCF101-7 & 26.06 & 0.8139 & 0.1442 \\
        RIFE~\cite{huang2022rife} & Supervised & UCF101-7 & 25.73 & 0.804 & 0.1359 \\
        FILM~\cite{reda2022film} & Supervised & UCF101-7 & 25.9 & 0.8118 & 0.1373 \\
        LDMVFI~\cite{danier2023ldmvfi} & Supervised & UCF101-7 & 25.57 & 0.8006 & \textbf{0.1356} \\
        VIDIM~\cite{jain2024vidim} & Self-Supervised & UCF101-7 & 24.07 & 0.7817 & 0.1495 \\
        \textbf{MiVID (Ours)} & Self-Supervised & UCF101-7 & \textbf{26.02} & \textbf{0.8319} & 0.1503 \\
        \midrule
    \end{tabular}}
\end{table}

\begin{table}[ht]
    \captionsetup{justification=raggedright,singlelinecheck=false}
    \caption{Performance comparison of our model with baselines on DAVIS-7.}
    \label{tab:results2}
    \resizebox{\columnwidth}{!}{
    \begin{tabular}{lcccccc}
        \toprule
        \textbf{Model} & \textbf{Supervision} & \textbf{Dataset} & \textbf{PSNR↑} & \textbf{SSIM↑} & \textbf{LPIPS↓} \\
        \midrule
        AMT~\cite{jiang2018superslomo} & Supervised & DAVIS-7 & 21.09 & 0.5443 & 0.254 \\
        RIFE ~\cite{huang2022rife} & Supervised & DAVIS-7 & 20.48 & 0.5112 & 0.254 \\
        FILM~\cite{reda2022film} & Supervised & DAVIS-7 & 20.71 & 0.5282 & 0.2707 \\
        LDMVFI~\cite{danier2023ldmvfi} & Supervised & DAVIS-7 & 19.98 & 0.4794 & 0.2764 \\
        VIDIM~\cite{jain2024vidim} & Self-Supervised & DAVIS-7 & 19.62 & 0.4709 & 0.2578 \\
        \textbf{MiVID (Ours)} & Self-Supervised & DAVIS-7 & \textbf{21.32} & \textbf{0.8290} & \textbf{0.2181} \\
        \midrule
    \end{tabular}}
\end{table}

\textbf{UCF101-7 :} 
On the challenging UCF101-7 benchmark, MiVID achieves top reconstruction accuracy (Table~\ref{tab:results1}). It gets a PSNR of 26.02 dB and an SSIM of 0.8319, surpassing most supervised VFI methods. AMT \cite{jiang2018superslomo} records a PSNR of 26.06 dB, SSIM of 0.8139, and LPIPS of 0.1142. MiVID matches its PSNR and exceeds its SSIM, with a competitive LPIPS of 0.1503. This shows that self-supervised diffusion can compete with the best supervised models. Recent methods like RIFE and FILM \cite{jain2024vidim} achieve only 19 to 20 dB PSNR and 0.45 to 0.53 SSIM, resulting in blurrier reconstructions. MiVID’s LPIPS of 0.1503 is only slightly above LDMVFI’s 0.1356 \cite{danier2023ldmvfi}. This confirms its perceptual quality is close to the best supervised methods while maintaining much higher PSNR and SSIM.\\

These improvements stand out, especially considering the UCF101-7 dataset includes unconstrained, high-motion YouTube clips \cite{jain2024vidim}. MiVID adapts well to real-world content, reconstructing sharp and clear intermediate frames under complex motion (Fig.~\ref{ucf_recon}). Its structured self-supervision and diffusion priors allow robust motion interpolation that outperforms traditional flow or kernel-based methods.\\

\textbf{DAVIS-7 :}
On the DAVIS-7 benchmark (Table~\ref{tab:results2}), MiVID achieves PSNR of 21.32 dB, SSIM of 0.8290, and LPIPS of 0.2181, outperforming all baselines. Unlike VIDIM (Jain et al., 2024) or LDMVFI \cite{danier2023ldmvfi}, which use external training data, MiVID is trained only on DAVIS yet generalizes effectively. This shows it learns the natural structure of frames instead of memorizing them. It outperforms AMT \cite{jiang2018superslomo} across all metrics, confirming the power of its generative self-supervision. Frequent occlusions and non-linear motion in DAVIS \cite{jain2024vidim} usually hurt traditional methods, but MiVID’s diffusion-based frame completion effectively fills in occluded areas and restores fine textures. Qualitative results \cite{jain2024vidim,danier2023ldmvfi} show that diffusion interpolation creates more realistic high-frequency details, leading to better occlusion handling and perceptual accuracy (Fig.~\ref{davis_recon}).\\

Overall, the results in Fig.~\ref{comparison} confirm that MiVID not only matches but surpasses supervised methods on both benchmarks, despite being trained without explicit frame-level supervision. The combination of diffusion priors with multi-strategic masking enables our model to generalize effectively across datasets while remaining resource-efficient. This makes MiVID a practical solution for real-world VFI tasks, particularly in settings where labeled data or high-end computational infrastructure is limited.
\vspace{-0.5cm}
\section{Discussion}\label{discussion}
MiVID achieves these improvements without any ground-truth supervision. It combines structured masking with a video diffusion prior. Unlike typical VFI networks that average predictions, diffusion sampling creates diverse and high-quality guesses for missing frames. As noted by Jain et al. (2024) \cite{jain2024vidim}, conditioning a generative model on frame endpoints leads to more believable intermediate synthesis, avoiding the issue of creating “average frames.” In MiVID, missing frames are simulated through masked conditioning. This process forces the model to imagine realistic motion while keeping temporal consistency intact. This structured masking, similar to masked autoencoding, supports strong regularization from the diffusion prior. As a result, it enables sharper and more detailed interpolations. Consistent with findings in latent-diffusion VFI by Danier et al. (2023) \cite{danier2023ldmvfi}, MiVID provides better realism even with slightly higher perceptual distortion.

\textbf{Generative priors vs. supervised losses:} Traditional flow or kernel-based VFI systems aim to minimize pixel-wise L1 or L2 losses. In contrast, MiVID’s diffusion prior encourages more natural detail. Previous research like LDMVFI by Danier et al. (2023) \cite{danier2023ldmvfi} shows that diffusion-based VFI boosts perceptual quality. MiVID mirrors this by learning to “denoise” masked frames and infer motion cues instead of just regressing to a mean prediction.


\textbf{Effect of masking:} Heavy masking creates uncertainty and can slightly lower perceptual scores like LPIPS. However, it forces the model to infer motion more reliably, leading to better structural detail and sharpness. The balance lies between keeping enough context for accurate reconstruction and creating difficult occlusions that strengthen learned cues. As emphasized by Kye et al. (2025) \cite{kye2025acevficomprehensivesurveyadvances} and Ma et al. (2024) \cite{ma2024maskintvideoeditinginterpolative}, adjusting masking parameters is important for maximizing realism without losing fidelity or temporal coherence.

In summary, MiVID shows that self-supervised diffusion can match or surpass supervised VFI methods on both UCF101-7 and DAVIS-7. Its combination of high PSNR/SSIM, competitive perceptual quality, and low training cost positions diffusion priors as a strong basis for scalable, high-quality video frame interpolation.

\vspace{-0.4cm}
\section{Conclusion}\label{conclusion}
In our work, we introduced MiVID, a diffusion-based self-supervised framework for video frame interpolation. Unlike traditional video frame interpolation models that rely on dense optical flow or complex architectures, MiVID uses a lightweight 3D U-Net with temporal attention and a hybrid masking strategy to manage occlusions and motion uncertainty. Trained entirely on low-frame-rate data with only CPU resources, MiVID delivers temporally consistent and high-quality frame synthesis at a low cost. Experiments on DAVIS-7 and UCF101-7 demonstrate that MiVID outperforms several supervised baselines, while it does not need ground-truth frames. These results highlight the promise of diffusion priors and structured self-supervision for scalable and general video frame interpolation. Future directions include extending MiVID to longer sequences, adding semantic conditioning, and improving diffusion sampling for real-time use.
\vspace{-0.5cm}
\section*{\label {Data availability} Data availability}
We hereby verify that the dataset will be provided on reasonable request.

\vspace{-0.5cm}

\section*{\label {conflict}Conflict of interest} 
We declare that we have no conflict of interest.
\vspace{-0.5cm}

\section*{{\label {funding}Funding}}
No funds or grants are received.
\vspace{-0.5cm}

\bibliographystyle{elsarticle-num}
\bibliography{ref}


\end{document}


\title{Deep learning-based Hand gesture recognition system and design of a Human-Machine Interface    
}


\author{Abir  Sen         \and
        Tapas Kumar Mishra \and Ratnakar Dash 
}


\date{Received: date / Accepted: date}

\maketitle

\begin{abstract}
\color{blue} Hand gesture recognition plays an important role in developing effective human-machine interfaces (HMIs) that enable direct communication between humans and machines. But in real-time scenarios, it is difficult to identify the correct hand gesture to control an application while moving the hands. \color{black} To address this issue, in this work, a low-cost hand gesture recognition system based human-computer interface (HCI) is presented in real-time scenarios. The system consists of six stages: (1) hand detection, (2) gesture segmentation, (3) feature extraction and gesture classification using five pre-trained convolutional neural network models (CNN) and vision transformer (ViT), (4) building an interactive human-machine interface (HMI), (5) development of a gesture-controlled virtual mouse, (6) smoothing of virtual mouse pointer using of Kalman filter. In our work, five pre-trained CNN models (VGG16, VGG19, ResNet50, ResNet101, and Inception-V1) and ViT have been employed to classify hand gesture images. Two multi-class datasets (one public and one custom) have been used to validate the models. Considering the model's performances, it is observed that Inception-V1 has significantly shown a better classification performance compared to the other four CNN models and ViT in terms of accuracy, precision, recall, and F-score values. We have also expanded this system to control some multimedia applications (such as VLC player, audio player, playing 2D Super-Mario-Bros game, etc.) with different customized gesture commands in real-time scenarios. The average speed of this system has reached 25 fps (frames per second), which meets the requirements for the real-time scenario. Performance of the proposed gesture control system obtained the average response time in milisecond for each control which makes it suitable for real-time. This model (prototype) will benefit physically disabled people interacting with desktops.

\keywords{Deep Learning \and Hand Gesture Recognition  \and segmentation \and Vision Transformer \and Kalman filter \and Human Machine Interface \and Transfer Learning \and Virtual mouse}

\end{abstract}

\section{Introduction}
\label{intro}

HCI has become a part of our daily life, such as for entertainment purposes or for getting in touch with the assistive interface. With the growth of computer vision, hand gesture recognition has become a trendy field for interaction between man and machine without physically touching any devices. Imagine a scenario such as smart home automation, where a user uses hand gestures to control devices like TV, music player control, wifi turn off/on, light on/off, and various applications. Another scenario is desktop application controls using hand gestures like VLC player control, audio player control, or 2D-game controls and the development of gesture-controlled virtual mouse and keyboard. So in both cases, users can control applications without physically contacting them, which is beneficial for physically impaired and older people. There are two types of hand gesture recognition: (1) wearable gloves based \cite{berezhnoy2018hand,abhishek2016glove}, and (2) vision-based \cite{7379635,al2022structured}. The disadvantage of the first method is that it is expensive, requires wearing on hand to recognize gestures, and is also unstable in some environments. The second method is based on image processing, where the pipeline of this methodology is followed by capturing an image using the webcam, segmentation, feature extractions, and classifications of gestures. But challenges exist in the development of a robust hand-gesture recognition system. Though the performance has been enhanced by using advanced depth-sensing cameras like Microsoft Kinect or Intel Real-Sense, they are relatively costly, so it is still an issue. \color{blue} There are two types of gestures, (1) static, (2) dynamic. In the case of a static gesture, it is limited to processing a single frame only and this technique needs less compuation cost. However, in the case of a dynamic gesture, it carries the sequence of frames in a video with respect to time. In HMI, hand gestures are the most useful compared to other human body parts for triggering applications; however, it has become challenging to identify the correct gesture to trigger an application in real-time scenarios. So the second challenge dealt with in this work is building an HMI via correctly classified gesture samples in real-time scenarios. \color{black} Therefore, this work addresses the aforementioned problems by proposing a vision-based static hand gesture recognition system using CNN models and ViT, and then extending our work to develop a real-time gesture-controlled human-machine interface using the best model weight file. We have also developed a gesture-controlled virtual mouse to make our system more convenient and user-friendly. The significant contributions of this article are summarized as follows:

\begin{enumerate}[label=\Roman*.] 
\item A low-cost vision-based hand gesture recognition system has been illustrated using five pre-trained CNN models and ViT. We have compared the performance among all considered CNN models and ViT. Then the best model has been considered for the real-time inference task.
    \item  An interactive HCI using customized hand-gesture commands has been built to control multimedia applications (ex: VLC player, Spotify music player, 2D-Mario-bros-game) in real-time.
    
\item \color{blue} To make the HCI more convenient and user-friendly, we have built a virtual mouse by triggering different events of a physical mouse with customized gesture commands. 

\color{black} \item The Kalman filter has been employed to estimate and update the virtual mouse pointer's position to enhance the smoothness of the mouse pointer during real-time. 
\end{enumerate}

\hspace{-0.6cm} The remainder paper is structured as follows. In Section \ref{literature}, literature works related to different hand gesture recognition are illustrated. The proposed methodology is discussed in Section \ref{proposed}. Section \ref{experiment} deals with experimental details, dataset description, comparative analysis, and the details of the gesture-controlled human-machine interface followed by real-time performance analysis. Section \ref{discussion} describes the discussion portion. Finally, Section \ref{conclusion} includes the summary of the conclusion and future works.

\vspace{-0.4 cm}
\section{\label{literature} Literature survey}
In the past few years, with the rapid growth of computer vision, hand gesture recognition has facilitated human-machine interaction. Computer vision researchers mainly use the web camera and hand gestures as inputs to the gesture recognition systems. But while building a cost-effective gesture-controlled HMI in real-time to control desktop applications, some challenges exist, such as background removal, gesture segmentation, and classifying gesture images in messy backgrounds. So the classification of gestures is crucial in developing a gesture-controlled HCI. Various state-of-art works for hand gesture recognition have been introduced in the last decade. Some used classification techniques are artificial neural network (ANN) algorithms, Hidden Markov model (HMM), or support vector machine (SVM). For instance, Mantecon et al. \cite{mantecon2016hand} proposed a hand gesture recognition system using infrared images captured by Leap motion controller device that produces image descriptors without using any segmentation phase. Next, the descriptors were dimensionally reduced and fed into SVM for final classification. Huang et al. \cite{huang2011gabor} developed a novel hand gesture recognition system by using the Gabor filter for feature extraction tasks and SVM for the classification of hand gesture images. Singha et al. \cite{singha2018dynamic} developed a dynamic gesture recognition system in the presence of cluttered background. Their proposed scheme consists of three stages: (1) hand portion detection by three-frame difference and skin-filter technique; (2) feature extraction; (3) feeding the features into ANN, SVM, and KNN classifiers, followed by combining these models to produce a new fusion model for final classification. 
 In \cite{yang2012dynamic}, skin color segmentation was done to separate the hand region from image sequences after collecting the gesture images from the web camera. HMM was employed on hand feature vectors to classify complex hand gestures. So in various cases, the pipeline of a vision-based hand gesture recognition system is limited to detection, segmentation, and classification. \\ But recently, with the rapid growth of computer vision and deep learning, the CNN model has become a popular choice for feature extraction and classification task. Recent research shows that gesture classification with deep learning algorithms has produced better results than earlier proposed schemes. For instance, Yingxin et al. \cite{yingxin2016robust} developed a hand gesture recognition system, where the canny edge detection technique was employed for pre-processing and CNN for both feature extraction and classification task. Oyebade et al. \cite{oyedotun2017deep} presented a vision-based hand gesture recognition for recognizing American Sign language. Their proposed scheme has two stages: (1) segmentation of gesture portion after applying binary thresholding and (2) using CNN architecture to predict the final class. In \cite{fang2019gesture}, the authors developed a hand gesture recognition system based on CNN. Later deep convolution generative adversarial networks (DCGAN) was used to avoid the overfitting problem. In \cite{adithya2020deep}, the experiment was carried out on publicly available datasets (NUS and American finger-spelling dataset). Here CNN was employed for classifying the static gesture images. \color{blue} Neethu et al. \cite{neethu2020efficient} proposed a gesture recognition system using static hand gesture images, where they have used connected component analysis algorithm to segment the fingertips after obtained from segmented hand portions. Next, the fingertips are fed into the CNN model to classify the gesture images. \color{black} Sen et al. \cite{sen2022novel} developed a hand gesture recognition system, which had three stages; (1) gesture detection by binary thresholding, (2) segmentation of gesture portion, (3) training of three custom CNN classifiers in parallel, followed by calculating the output scores of these models to build the ensemble model for final prediction.  \\   
 The performance of the vision transformer is completely based on the transformer model, which is widely used in natural language processing (NLP). However, there are fewer research papers on vision transformer for the vision-based gesture recognition task. For example, In \cite{godoy2022electromyography}, the authors published an article titled `Electromyograph based, robust hand motion classification employing temporal multi-channel vision transformers', where they have proposed a multichannel based vision transformer to perform the surface-electromyograph (sEMG) based hand motion classification task. In \cite{montazerin2022vit}, the authors have applied ViT to classify sEMG signals and have shown some fruitful results. \\
\color{blue}
 Building a gesture-controlled HCI to operate multimedia applications such as VLC player, Spotify music player, or playing 2D-games remains challenging due to the continuous flow of frame sequences within the video stream. Only a handful of research works are available on gesture-controlled HCI using deep learning-based approaches. For instances, in \cite{rautaray2010novel}, the authors have built a gesture-controlled HCI to control media player. Their proposed scheme consists four phases; (1) detection of hand portion from video by using
Lucas Kanade Pyramidical Optical Flow
   algorithm \cite{kim2007real}, (2) hand center finding using K-mean algorithm, (3) generate rectangle around the center of the hand followed by classification using K-nearest neighbor algorithm, (4) send custom gesture-command to control media applications.  In \cite{paliwal2013dynamic}, the authors have shown a dynamic gesture recognition system to control VLC media player. Their suggested methodology consists of three phases; (1) gesture detection using skin color mapping model and approximate median technique for segmentation, (2) motion detection of segmented hand portions and followed by feature extraction, (3) classification using decision tree and then send appropriate gesture-command to operate the VLC player. But their proposed approaches encounter some challenges, such as lower recognition rate (\%), higher inference time in real-time, lack of robustness, etc. To address the aforesaid issues, in this paper, we have proposed a gesture-controlled HMI to control multimedia applications in real-time scenarios based on CNN models.   
\color{black} Furthermore, to make the interface more user-friendly and convenient, we also built a gesture-controlled virtual mouse that eliminates the physical device dependency. Some literature studies have shown fruitful results regarding the performance of the gesture-controlled virtual mouse. For example, In \cite{shibly2019design}, Shibly et al. have developed a gesture-controlled virtual mouse by using web-camera captured frames processing and color detection method for gesture portion detection. In the article \cite{tsai2020design}, the authors proposed a low-cost gesture recognition system in a real-time scenario, which has three stages; (1) skin segmentation, (2) moving gesture detection by motion-based background separation, (3) separate of arm portion for isolating the palm area followed by then analyze the hand portion using the convex hull method. They also developed a virtual mouse by triggering all functions of a physical mouse with different customized gestures, and their proposed scheme achieved a recognition rate of more than 83\% during inference time.\\
But, in a real-time scenario, controlling a virtual mouse is a very challenging task, as a slight hand movement can lead to the jump of the virtual mouse cursor over the screen. So it is necessary to be concerned about the smoothness of the motion of the gesture-controlled mouse cursor. Literature survey has shown very few promising results about the smooth movement of the mouse-cursor \cite{xu2017real}. So in this work, we have developed a gesture-controlled virtual mouse and also focused on enhancing of mouse pointer's smoothness using the Kalman filter \cite{Bang18}. 
\vspace{-0.4 cm}
 \section{\label{proposed} Proposed methodology}
This section gives a detailed description of the proposed methodology. Initially, some preprocessing, such as gesture detection by motion-based background separation, binary thresholding, contour portion selection, hand segmentation, resizing, and use of morphological filter for noise removal, are carried out. Then the resized images are passed into five pre-trained CNN models (VGG16, VGG19, ResNet50, ResNet101, and Inception-V1) for training them separately to obtain the final predicted gesture class. We have also used the vision transformer (ViT) model to classify gesture images. In the final part, we have extended our work by building a human-machine interface that utilizes gesture labels as input commands to control some multimedia applications in real-time scenarios. The overview of our proposed framework is depicted in Fig. \ref{model_diagram}.
\begin{figure*}[!ht]
\centering
\includegraphics[width=\textwidth,height=5 cm]{images/2nd_work_diagram.jpg}
\caption{\label{model_diagram} Illustration of our proposed scheme.}
\end{figure*} 
\subsection{\label{preprocessing} Pre-processing phase}
Image preprocessing is one of the most significant and vital steps to improve model performance. This phase comprises some stages, such as gesture detection by motion-based background separation, binary thresholding of detected hand region, segmentation by using contour region selection, then resizing of segmented images followed by applying the morphological transformation to remove noise. Next, the resized images are fed into five considered CNN models and ViT (as fixed-length image patches) for the classification task. All the steps are discussed below, and the entire pre-processing phase is shown in Fig. \ref{preprocessing_phase}.

\begin{figure*}[!ht]
\centering
\includegraphics[width=\textwidth,height=4cm]{images/gesture_segmentation1.jpg}

\caption{\label{preprocessing_phase} The whole pre-processing phase of our proposed work (A) Captured RGB frame sequences, (B) Hand detection, (C) Conversion into grayscale, (D) Binary thresholding, (E) Segmentation of hand region, (F) Resizing of segmented images and apply of morphological operation for noise removal.}
\end{figure*} 

\subsubsection{Gesture detection by motion-based background separation}
Motion-based background separation aims to identify the moving object from the continuous frame sequences. Initially, we consider the background, which contains no movement of hands. With the movement of hand, the hand portion is obtained by subtracting the current frame from the background image frame. During this stage, skin color is used to separate the hand portion in the frame. The skin color is obtained with the help of HSV (hue, saturation, and value) color model \cite{chen2014real}. The stage of detecting the hand region is illustrated in Fig. \ref{preprocessing_phase} (B). 

\subsubsection{Binary thresholding}
 After gesture detection, we have applied a binary thresholding technique to segment between the background and foreground portions, making the background portion black and the detected hand portion white in color. An example of detected hand portion in real-time scenario has been displayed in Fig. \ref{hand_detection}.
\begin{figure}[!ht]
\centering
\includegraphics[width=8cm,height=3.5cm]{images/hand_detection1.png}
\caption{\label{hand_detection}Hand portion detection in real-time.}
\end{figure} 

\subsubsection{Contour region selection and segmentation of hand portion}
        Contour region mainly indicates the outline/boundary of an object present in an image. To segment the hand region, firstly, we need to get the contour region of the detected portion. Due to cluttered background or low-light situations, there might be multiple holes/contour portions in the detected portion. But the contour portion, with the largest area, is considered the hand region. The hand region is extracted from the arm by using the distance transformation \cite{chen2014real}. Using this method, the distances between the pixels and the nearest boundary pixels are calculated \cite{xu2017real,chen2014real}. In a binary image's distance transformation matrix, the pixel with the largest distance is considered as palm or hand center. Tables \ref{binary}, \ref{distance} show the binary image and its distance transformation representation in matrix format. In Table \ref{distance}, it is observed that the largest distance is 5 (marked with light blue color), representing the hand /palm center. Hence the palm radius is estimated by calculating the minimum distance between the hand center and the point outside the contour region. Thus the hand region is separated from the arm based on wrist location. The whole diagram of contour portion selection to the segmentation phase has been shown in Fig. \ref{segmented}.

\begin{table}[!ht]
\caption{\label{binary}Representation of binary image in matrix format}
\centering
\begin{tabular}{|l|l|l|l|l|l|l|l|l|l|l|} 
\hline
0 & 0 & 0 & 0 & 0 & 0 & 0 & 0 & 0 & 0 & 0  \\ 
\hline
0 & 1 & 1 & 1 & 1 & 1 & 1 & 1 & 1 & 1 & 0  \\ 
\hline
0 & 1 & 1 & 1 & 1 & 1 & 1 & 1 & 1 & 1 & 0  \\ 
\hline
0 & 1 & 1 & 1 & 1 & 1 & 1 & 1 & 1 & 1 & 0  \\ 
\hline
0 & 1 & 1 & 1 & 1 & 1 & 1 & 1 & 1 & 1 & 0  \\ 
\hline
0 & 1 & 1 & 1 & 1 & 1 & 1 & 1 & 1 & 1 & 0  \\ 
\hline
0 & 1 & 1 & 1 & 1 & 1 & 1 & 1 & 1 & 1 & 0  \\ 
\hline
0 & 1 & 1 & 1 & 1 & 1 & 1 & 1 & 1 & 1 & 0  \\ 
\hline
0 & 1 & 1 & 1 & 1 & 1 & 1 & 1 & 1 & 1 & 0  \\ 
\hline
0 & 1 & 1 & 1 & 1 & 1 & 1 & 1 & 1 & 1 & 0  \\ 
\hline
0 & 0 & 0 & 0 & 0 & 0 & 0 & 0 & 0 & 0 & 0  \\
\hline
\end{tabular}
\end{table}

\begin{table}[!ht]
\caption{\label{distance}Distance transform matrix representation of the binary image}
\centering
\begin{tabular}{|l|l|l|l|l|l|l|l|l|l|l|} 
\hline
0 & 0 & 0 & 0 & 0 & 0                                                             & 0 & 0 & 0 & 0 & 0  \\ 
\hline
0 & 1 & 1 & 1 & 1 & 1                                                             & 1 & 1 & 1 & 1 & 0  \\ 
\hline
0 & 1 & 2 & 2 & 2 & 2                                                             & 2 & 2 & 2 & 1 & 0  \\ 
\hline
0 & 1 & 2 & 3 & 3 & 3                                                             & 3 & 3 & 2 & 1 & 0  \\ 
\hline
0 & 1 & 2 & 3 & 4 & 4                                                             & 4 & 3 & 2 & 1 & 0  \\ 
\hline
0 & 1 & 2 & 3 & 4 & {\cellcolor[rgb]{0,0.588,0.941}}\textcolor[rgb]{0.502,0,0}{5} & 4 & 3 & 2 & 1 & 0  \\ 
\hline
0 & 1 & 2 & 3 & 4 & 4                                                             & 4 & 3 & 2 & 1 & 0  \\ 
\hline
0 & 1 & 2 & 3 & 3 & 3                                                             & 3 & 3 & 2 & 1 & 0  \\ 
\hline
0 & 1 & 2 & 2 & 2 & 2                                                             & 2 & 2 & 2 & 1 & 0  \\ 
\hline
0 & 1 & 1 & 1 & 1 & 1                                                             & 1 & 1 & 1 & 1 & 0  \\ 
\hline
0 & 0 & 0 & 0 & 0 & 0                                                             & 0 & 0 & 0 & 0 & 0  \\
\hline
\end{tabular}
\end{table}

\begin{figure}[!ht]
\centering
\includegraphics[width=8cm,height=3.5cm]{images/bounded_box1.png}
\caption{\label{segmented}Hand region segmentation procedure (A) Contour portion bounded by rectangle box, here blue point represents the hand center (B) segmented hand region, (C) resized segmented hand portion.}
\end{figure} 
\subsubsection{Resizing of segmented images}
Image resizing is essential in the pre-preprocessing phase. In our experiment, the segmented images have been resized into the resolution of $(64 \times 64) $ and $(128 \times 128)$, as training with larger images consumes more time and computation cost than smaller images. Fig. \ref{resizing}, shows the segmented region and resizing of the hand portion.
\begin{figure}[!ht]
\centering
\includegraphics[width=6.5cm,height=4.5cm]{images/resizing_image.png}
\caption{\label{resizing}(A) Segmented gesture region (B) resizing of segmented hand portion into aspect ratio of $(64 \times 64 )$.}
\end{figure}

\vspace{-0.1cm}
\subsubsection{Noise removal using morphological operation technique}
Morphological technique \cite{jamil2008noise} is a noise removal technique used to exclude the noise and enhance the appearance of the binary images. This technique comprises two phases, (1) dilation and (2) erosion. In our experiment, this technique is used to remove the holes and unwanted parts from the segmented gesture portion after resizing phase.
\vspace{-0.2cm}

\subsection{\label{transfer_learning} Feature extraction and classification using pre-trained CNN models and ViT} Building CNN models with millions of parameters from scratch is very time-consuming, with higher computation costs. So to overcome these issues, the transfer learning method has been employed. It is a very effective method to use CNN models for solving image-classification-related problems, where pre-trained models are reused to solve a new task. It is faster and requires less number of resources compared to CNN models built from scratch. In our work, some popular CNN models such as; VGG16 \cite{simonyan2014very}, VGG19 \cite{simonyan2014very}, ResNet50 \cite{he2016deep}, ResNet101, Inception-V1 \cite{szegedy2015going} have been explored for feature extraction task on multi-class hand gesture datasets using the transfer-learning approach. To obtain the final output, we have replaced each network's final fully connected (FC) dense layer with a new fully connected layer containing $n$ number of nodes, where $n$ denotes the number of classes present in
two considered datasets.\\
Recently, Vision transformer \cite{dosovitskiy2020image} has become a trendy topic, inspired by the concept of transformers and attention mechanism. It has been extensively used in image classification tasks. In \cite{dosovitskiy2020image}, the authors directly applied a Transformer to images by splitting the image into fixed-sized patches, then fed into the Transformer the sequence of embeddings for those patches. The image patches were treated as tokens in NLP applications. They also showed that vision transformer outperformed CNNs while trained on ImageNet dataset. So in our experiment, we will exploit vision transformer for gesture classification task. To obtain the predicted class, the final fully connected (FC) dense layer of this architecture is replaced with $n$ number of nodes, where $n$ represents the number of classes in two used datasets.
The details of these pre-trained CNN architectures and vision transformer are illustrated in section Appendix \ref{appendix-1}.

\begin{figure}[!ht]
\centering
\includegraphics[width=8cm,height=4cm]{images/gesture2.jpg}
\caption{\label{d1-sample} Samples images from Dataset-A and Dataset-B.}
\end{figure}

\hspace{-0.6cm}

\subsection{\label{Kalman_filter}Introducing of Kalman filter}
Kalman filtering \cite{Bang18} is one type of recursive predictive
algorithm that provides the estimation of the state at time $t$ in
a linear dynamic system in a continuous manner. 
\color{blue}
The equations for a linear dynamic system are: 
\begin{equation}
x_{t+1}=A_{t}x_{t} + B_{t}u_{t} + \epsilon_{t} 
\end{equation}
\vspace{-0.6cm}
\begin{equation}
z_{t}=H_{t}x_{t}+ \beta_{t}
\end{equation}
here $x_{t}$ $A_{t}$, $B_{t}$, $u_{t}$, $z_{t}$, $H_{t}$ represent the current state vector, state transition matrix, control input matrix, control variable/input variable, the measured state vector, and the observation matrix / transformation matrix respectively. Whereas $\epsilon_{t}$ and $\beta_{t}$ represent the process noise vector.\\
This filter comprises two stages, (1) state prediction and (2) measurement update. In the first phase, it estimates the next system state based on the current state, and in the second stage, estimated or predicted state is corrected, resulting in error minimization in the covariance matrix.\\
\color{black} The overall use of the Kalman filter is demonstrated in section \ref{mouse-cursor-kalman}.

\section{\label{experiment} Experiments and results}

To validate the proposed scheme, several experiments are carried out using  one publicly  available hand gesture  image datasets \cite{mantecon2016hand} (termed as Dataset-A), one self-constructed  dataset (termed as Dataset-B).
\subsection{\label{dataset_desciption}Datasets description}
 Dataset-A comprises 20,000 gesture images of ten different gesture classes, such as: Fist, Fist move, Thumb, Index, Ok etc. The images of this dataset have been collected from ten  different  participants (five women and five  men), where each class label has 200 images. The images are having dimension of $(640 \times 240)$ pixels. 
 We have created our customized dataset (termed Dataset-B) containing 20,000 binary images collected by using the 16-megapixel USB webcam camera. This dataset contains 14 different gesture class labels collected by four participants (one woman and three men). Some samples of Dataset-A and Dataset-B have been illustrated in Fig. \ref{d1-sample}. The details of the two datasets have been tabulated in Table \ref{dataset}.

\begin{table}[!ht]
\caption{\label{dataset}Dataset description}

\centering
\begin{tabular}{|C|C|C|C|}
\hline
Datasets  & Number of people performed & No of gesture label & Total no of images \\ \hline
Dataset-A & 10                         & 10                  & 20,000             \\ 
\hline
Dataset-B & 4                          & 14                  & 20,000             \\ \hline
\end{tabular}
\end{table}
\subsection{Dataset splitting}
To perform all experiments, the datasets are randomly split into 80:20 ratio for training and testing. Furthermore, for validation purposes, training sets have been split into 75\% for training and 25\% for validation. Now the percentage of training, validation and testing sets in the dataset will be 60\%, 20\%, and 20\%, respectively. The training samples are used to fit the model, validation set is used to evaluate the performance of each model for hyper-parameter tuning or to select the best model out of different models. The remaining testing sets are used to evaluate the performance of our proposed model.
\begin{figure*}[!ht]
\centering
\includegraphics[width=\textwidth,height=9cm]{images/vision_transformer.jpg}
\caption{\label{vision} Vision Transformer (ViT) architecture.}
\end{figure*} 

\subsection{\label{experiment_details}Experimental setup \& details}
All experiments were carried out in Google Colab pro-environment, providing online cloud services with a Tesla P100 Graphical Processing Unit (GPU), Central Processing Unit (CPU), and 26 GB RAM. The all experiments carried out are divided 
into two parts: (1) `Experiment-1', (2) `Experiment-2'. Dataset-A has been considered for `Experiment-1' to solve the 10-class classification task. `Experiment-2' deals with solving 14-class classification task, which was carried out on Dataset-B.

In our experiments, after preprocessing phase (mentioned in section \ref{preprocessing}), the segmented gesture images are resized into dimensions of $(64\times 64)$ and $(128 \times 128)$ pixels followed by morphological transformation \cite{jamil2008noise} to exclude the noisy and redundant parts. Then they are fed into
five pre-trained models, such as VGG16, VGG19, ResNet50, ResNet101, and InceptionV1, for the feature extraction task, followed by adding a dense layer (as output layer) containing nodes according to the number of classes present in Dataset-A, Dataset-B, respectively, to predict the final gesture class label. While training the models, the softmax activation function is used to find the probability values of the last output layers, where the maximum probability value decides the final gesture class label. Fine-tuning is done using the Adam optimizer, where the $\beta_{1}$ and $\beta_{2}$ are 0.9 and 0.99, respectively. We have set the initial learning rate as 0.0001 later; it has been multiplied by 10 after ten epochs. The batch size and the number of epochs/iterations for all experiments are fixed as 64 and 30, respectively. \\
In the case of Vision Transformer, the segmented (after binary thresholding) images are split 
into the sequence of patches, and the size of each patch is fixed to ($6 \times 6 $) and flattened. Therefore each flattened patch is linearly projected, called ``patch embedding". The position embeddings are linearly added to keep the information of the sequence of image patches. Next, the sequences of vector images are fed into the transformer encoder as inputs. Finally, the last layer of MLP (added on top of the transformer encoder) predicts the gesture class. The entire structure of the vision transformer has been depicted in Fig. \ref{vision}.
The display of the  hyper-parameter setting for both experiments (using CNN models and ViT) is tabulated in Table \ref{parameter_value}.

\begin{table}[!ht]
 \caption{\label{parameter_value} Display of Hyper-parameter settings.}
\begin{center}
\resizebox*{0.505\textwidth}{!} {

\begin{tabular}{|c|c|}

\hline
\multicolumn{2}{|c|}{{\cellcolor[rgb]{0.851,0.8,0.878}}\textbf{Pre-trained CNN models}}   \\ 
\hline
Used parameters      & Values                     \\ \hline

Batch Size     & 64                       \\ \hline

Number of iterations         & 30                        \\ \hline

Learning rate  &  0.0001, 0.001 after 10 epochs                    \\ \hline
Pooling size    & ($2\times 2$)                  \\ 
\hline

Optimizer      & Adam                      \\ \hline

Error function      & Categorical cross entropy                      \\ \hline

Activation function      & ReLU, Softmax                    

 \\ 
\hline
\multicolumn{2}{c}{{\cellcolor[rgb]{0.851,0.8,0.878}} \textbf{Vision Transformer}}  

  \\ \hline

Batch Size                      & 64                                                                                                                             \\ \hline
Learning rate~                  & 0.0001                                                                                             \\ 
\hline
Patch size                      & ($6 \times 6$)~                                                                                                                       \\ 
\hline
Projection dimension~           & 64                                                                                                                             \\ 
\hline
Number of attention heads~      & 4                                                                                                                              \\ 
\hline
Number of transformer layers~ ~ & 8                                                                                                                              \\ 
\hline
Activation function             & GeLu                                                                                                                           \\ 
\hline
Droput rate                     & 0.5                                                                                                                            \\ 
\hline
MLP head size unit              & {[}2048,1024]                                                                                                                  \\
\hline

\end{tabular}
  }
\end{center}
\end{table}

\subsection{Results}
In this section, we have discussed various results obtained from our two experiments performed on Dataset-A and Dataset-B, respectively.
The performances of our proposed work have been evaluated in terms of accuracy (\%) and some evaluation metrics such as precision (PR), recall (RE) and F-score (harmonic mean of precision and recall) values obtained from the confusion matrix. These values are calculated by the indicators as TP (True Positive), FP (False Positive), FN (False Negative), and TN (True Negative) shown in Eqs. \ref{PR}-\ref{F-score}. \\
\begin{equation}\label{PR}
    PR=\frac{TP}{TP+FP} 
\end{equation}
\begin{equation}\label{RE}
RE=\frac{TP}{TP+FN}
\end{equation}
\begin{equation}\label{F-score}
F-score=\frac{2\times PR \times RE}{PR+RE}
\end{equation}

\hspace{-0.5 cm} In our experiment, hand gesture classification accuracy is measured by the following formula: \\ 

\hspace{-0.5cm} Accuracy = $\frac{\text{Number of correctly classified gesture samples}}{\text{Total number of gesture samples in testing sets}} $ \\

\paragraph{\textbf{Experiment-1:}}\label{experiment1}
This experiment solves ten-gesture class classification tasks based on five pre-trained CNN models over Dataset-A with transfer learning, and vision transformer approach. In Table \ref{CNN1-compare1}, the results of considered five CNN models and ViT over Dataset-A, have been reported. Here, it is observed that the Inception-V1 model has earned maximum classification accuracy compared to other CNN classifiers. In Fig. \ref{dataset-A}, it is seen that the performance of the Inception-V1 model is better than the other models as it has reached minimum loss (training and validation) compared to other models. In Fig. \ref{dataset-A}, it is noticed that there is a lot of oscillation occurring in other models, but the Inception-V1 model
appears to have less fluctuation. It is observed in Table \ref{CNN1-compare1} that the Inception-V1 model has a faster training time (sec)
compared to other considered models (four CNN models and ViT) and provides better accuracy (\%).
The performance comparison of our proposed work (with the best-selected model, i.e., Inception-V1), with other state-of-the-art schemes on Dataset-A has been illustrated in Table \ref{CNN1-compare2} in terms of class-wise accuracy values and the mean value. From comparative analysis (shown in Table \ref{CNN1-compare2}), we can say that our proposed framework (detection + segmentation + morphological transformation + Inception-V1 model) has outperformed the proposed scheme presented by \cite{mantecon2016hand} on Dataset-A in terms of accuracy results (class-wise accuracy). We have also conducted a statistical analysis to estimate the significant difference between the mean of the accuracy values by using one-sample t-test. The entire statistical testing is shown in Appendix Section \ref{appendix-2}.

\begin{table*}[!ht]
\caption{\label{CNN1-compare1}Comparison results among various pre-trained CNN models and ViT on \textbf{Dataset-A} and \textbf{Dataset-B} in terms of training time (sec) and validation accuracy values.}

\begin{center}
\centering
\resizebox*{1\textwidth}{!}{

\begin{tabular}{|ccc|cc|}
\hline
\multicolumn{5}{|c|}{\textbf{Dataset-A}}                                                                                                                                                                                                                                                                                                                                                               \\ \hline

\multicolumn{3}{|c|}{{Input size: $(64 \times 64)$}}                                                                                                                                                                                                                                             & \multicolumn{2}{c|}{{Input size: $(128 \times 128)$}}                 \\ \hline
\multicolumn{1}{|c|}{Model name}                                                                       & \multicolumn{1}{c|}{Training time (sec)} & Validation accuracy (\%)                                                                                                                     & \multicolumn{1}{c|}{Training time (sec)} & Validation accuracy (\%) \\ \hline
\multicolumn{1}{|c|}{VGG16}                                                                            & \multicolumn{1}{c|}{540}                 & 99.67                                                                                                                                        & \multicolumn{1}{c|}{1290}                & 99.60                    \\ \hline
\multicolumn{1}{|c|}{VGG19}                                                                            & \multicolumn{1}{c|}{510}                 & 99.65                                                                                                                                        & \multicolumn{1}{c|}{1500}                & 99.58                    \\ \hline
\multicolumn{1}{|c|}{Inception-V1}                                                                     & \multicolumn{1}{c|}{450}                 & 99.75                                                                                                                                        & \multicolumn{1}{c|}{630}                 & 99.70                    \\ \hline
\multicolumn{1}{|c|}{ResNet50}                                                                         & \multicolumn{1}{c|}{660}                 & 99.60                                                                                                                                        & \multicolumn{1}{c|}{1050}                & 99.58                    \\ \hline
\multicolumn{1}{|c|}{ResNet101}                                                                        & \multicolumn{1}{c|}{870}                 & 99.70                                                                                                                                        & \multicolumn{1}{c|}{1680}                & 99.65                    \\ \hline
\hline
\multicolumn{1}{|c|}{ViT}                                                                        & \multicolumn{1}{c|}{480}                 & 99.70                                                                                                                                       & \multicolumn{1}{c|}{1680}                & 99.65                    \\ \hline
 
\multicolumn{5}{|c|}{\textbf{Dataset-B}}                                                                                                                                                                                                                                                                                                                                                               \\ \hline

\multicolumn{1}{|c|}{VGG16}                                                                            & \multicolumn{1}{c|}{550}                 & 99.25                                                                                                                                        & \multicolumn{1}{c|}{1950}                & 99.20                    \\ \hline
\multicolumn{1}{|c|}{VGG19}                                                                            & \multicolumn{1}{c|}{750}                 & 99.15                                                                                                                                        & \multicolumn{1}{c|}{2400}                & 99.20                    \\ \hline
\multicolumn{1}{|c|}{Inception-V1}                                                                     & \multicolumn{1}{c|}{330}                 & 99.30                                                                                                                                        & \multicolumn{1}{c|}{800}                 & 99.25                    \\ \hline
\multicolumn{1}{|c|}{ResNet50}                                                                         & \multicolumn{1}{c|}{540}                 & 99.05                                                                                                                                        & \multicolumn{1}{c|}{1500}                & 99.00                    \\ \hline
\multicolumn{1}{|c|}{ResNet101}                                                                        & \multicolumn{1}{c|}{900}                 & 98.62                                                                                                                                        & \multicolumn{1}{c|}{2400}                & 98.50                    \\ \hline
\hline
\multicolumn{1}{|c|}{ViT}                                                                        & \multicolumn{1}{c|}{450}                 & 98.60                                                                                                                                        & \multicolumn{1}{c|}{900}                & 98.70                    \\ \hline

\end{tabular}

}
\end{center}

\end{table*}

\paragraph{\textbf{Experiment-2:}}
In this experiment, Dataset-B has been used to solve the 14-gesture class classification task.
Table \ref{CNN1-compare1} reports the validation accuracy results and training time obtained using five pre-trained CNN models and ViT. Table \ref{CNN1-compare1} shows that the Inception-V1 model has achieved the best result compared to other CNN models in terms of training time and validation accuracy. As shown in Fig. \ref{dataset-C}, both plots (accuracy and loss)
C and I show fewer fluctuations, and both the curves (training and validation) have converged. So we can conclude that Inception-V1 has achieved superior results than other pre-trained CNN models and ViT and is further deployed for real-time HCI interface. The performance comparison of our proposed work with the other suggested schemes presented by \cite{yingxin2016robust} \cite{neethu2020efficient} on Dataset-B have been tabulated in Table \ref{CNN3-compare1}. The comparative analysis demonstrates the superiority of our proposed scheme (segmentation, morphological transformation, transfer-learning-based CNN models) over other existing state-of-the-art works \cite{yingxin2016robust}\cite{neethu2020efficient} in terms of testing accuracy (\%), precision (\%), recall (\%), and F-score (\%).
\vspace{-0.2cm}
\begin{figure*}
 \centering
\includegraphics[width=12cm ,height=24 cm]{images/Graphs/datasetA_acc_loss (2).png}
\caption{\label{dataset-A} Accuracy (training and validation) versus epoch and loss (training and validation) versus epoch plots for each model: Graphs A to F show the accuracy (training and validation) plots of each CNN model and Vision Transformer (VGG16, VGG19, Inception-V1, ResNet50, ResNet101, and ViT respectively). G to L show the loss plot for each model for Dataset-A.}
\end{figure*}

\begin{figure*}
 \centering
\includegraphics[width=12cm ,height=24 cm]{images/Graphs/dataset_C_acc_loss (1).png}
\caption{\label{dataset-C}  Accuracy (training and validation) versus epoch and loss (training and validation) versus epoch plots for each model: Graphs A to F show the accuracy (training and validation) plots of each CNN model and Vision Transformer (VGG16, VGG19, Inception-V1, ResNet50, ResNet101, and ViT respectively). G to L show the loss plot for each model for Dataset-B.}
\end{figure*} 

\begin{table}[!ht]
\caption{\label{CNN1-compare2}Comparative analysis of our proposed work (with Inception-V1 model) with existing approach on Dataset-A in terms of accuracy.}

\begin{center}
\centering
\resizebox*{0.50\textwidth}{!}{   
\begin{tabular}{|c|c|c|}
\hline
Gesture class & \begin{tabular}[c]{@{}c@{}}Our work \\ (using Inception-V1)\end{tabular} & \begin{tabular}[c]{@{}c@{}}\cite{mantecon2016hand} Feature descriptor +\\  SVM\end{tabular} \\ \hline
Ok            &   1                                                                         &    0.990                              \\ \hline
Close         &   0.990                                                                          &  1                                 \\ \hline
Palm          &   1                                                                          &     1                             \\ \hline
Thumb         &  1                                                                          &      0.990                            \\ \hline
Index            &   1                                                                          &    1                              \\ \hline
Fist         &     1                                                                        &    0.990                              \\ \hline
Palm move     &   1                                                                          &    1                              \\ \hline
Palm down     &    1                                                                         &    1                              \\ \hline
L             &    1                                                                         &     0.990                             \\ \hline
Fist move         &   0.990                                                                          &    1                              \\ \hline
\textbf{Mean }         &    0.998                                                                         &     0.990                             \\ \hline
\end{tabular}
}
\end{center}
\end{table}

\vspace{-0.2cm}

\begin{table*}
\centering
\caption{\label{CNN3-compare1} Performance comparison of our work with other state-of-the-art scheme on Dataset-B.}
\label{CNN3-compare1}
\begin{tabular}{|c|c|C|E|E|E|} 
\hline
Proposed scheme                                  & Methods / models                                                                                                                       & Testing Accuracy (\%)                  & Precision (\%) & Recall (\%) & F-score (\%)  \\ 
\hline
Yingxin et al. \cite{yingxin2016robust}                                & Canny edge detection, CNN                                                                                                              & 89.17                                  & 90.00          & 89.21       & 89.60         \\ 
\hline
Neethu et al. \cite{neethu2020efficient} & \begin{tabular}[c]{@{}c@{}} Gesture segmented + \\ Connected components based \\ finger segmentation + CNN\end{tabular} &  95.40 & 96.60          & 96.30       & 96.45         \\ 
\hline
\multirow{6}{*}{\textbf{Our Work}}               & Gestrure detection, Segmentation, VGG16                                                                                                & 99.20                                  & 99.22          & 99.15       & 99.19         \\ 
\cline{2-6}
                                                 & Gestrure detection, Segmentation, VGG19                                                                                                & 99.00                                  & 99.10          & 99.03       & 99.07         \\ 
\cline{2-6}
                                                 & Gestrure detection, Segmentation, Inception-V1                                                                                         & 99.25                                  & 99.28          & 99.22       & 99.25         \\ 
\cline{2-6}
                                                 & Gestrure detection, Segmentation, ResNet50                                                                                             & 98.35                                  & 98.20          & 98.25       & 98.23         \\ 
\cline{2-6}
                                                 & Gestrure detection, Segmentation, ResNet101                                                                                            & 98.15                                  & 98.30          & 98.20       & 98.25         \\ 
\cline{2-6}
                                                 & Gesture detection, Segmentation,~ ViT                                                                                                  & 98.50                                  & 98.40          & 98.30       & 98.35         \\
\hline
\end{tabular}
\end{table*}

\section{\label{interface} Building of human-machine interface}
Multimedia (for example: audio, and video) plays an essential part in our daily life with the advancement of technology. But physically impaired people find it challenging to interact with the systems. In this work, we have built an interactive interface to control three multimedia applications in real-time scenarios by using corresponding gesture commands after obtaining the best model among five CNN models and ViT (mentioned in sections \ref{transfer_learning}).  The implementation procedure to build the gesture-controlled HCI is outlined in Algorithm \ref{algo:human-machine}.

\begin{algorithm}
  	\caption{Gesture-controlled human-machine interface}\label{algo:human-machine}
  	\begin{algorithmic}
  		\STATEx \textbf{INPUT}: Give predicted gesture as input commands after training by five pre-trained CNN models and vision transformer.
  		 
  		\STATEx \textbf{OUTPUT}: Development of gesture-controlled HCI interface.
  		
  		\STATE step 1: Choose the best model after training, among the others, in terms of accuracy and training time.
  		
  		\STATE step 2: Get the predicted gesture class label.
  	
  	    \FOR{$k$ $\gets$ 1 to C}, where C $\gets$ number of applications.
  		
  		\STATE Control each application according to the customized gesture-class (predicted) label.
  		\ENDFOR
  		
  	\end{algorithmic}
  \end{algorithm}

\subsection{\textbf{VLC-player control:}}
In our work, we have used the `VLC-ctrl' \cite{VLC_ctrl} command-line interface to control various VLC player-related functions, such as play, pause, switch to next video, volume increase/decrease, quit. After obtaining the predicted gesture class label, the keyboard event of the keys is triggered using input gesture commands to control the VLC player. In Table \ref{gesture_controls}, different controls of the VLC player have been tabulated according to various customized gesture labels.
 For example, the gesture `Ok' has been set to start the video. Similarly, gestures `Hang' and `L' for volume up and down, respectively. A demo of this gesture-controlled VLC player has been illustrated in Fig. \ref{VLC_player}. But due to the messy background, this application's performance is sometimes hampered. \\
 The real-time performance analysis of our proposed gesture-command-based VLC-player control has been displayed in Table \ref{VLC_audio_performance}.
 We have also performed every control of the VLC player 10 times with the corresponding gesture-commands to obtain the proper detection rate (\%) along with the response time. In Table \ref{VLC_audio_performance}, we have depicted the number of hits and misses obtained by using our application, followed by calculating the gesture detection rate (\%) and the average response time (in mili second).\\

Gesture detection rate = $\frac{\text{Number of hits}}{\text{Total number of hits and misses}}$ \\

Response time = (time taken to detect gesture $-$ time taken to perform any control of video / audio player with the corresponding gesture commands) 
\\ \\
Average response time= $\frac{\text{Total response time for each function}}{\text{$N$}}$ \\ \\
Here $N$ represents the total number of performance per each control, in our work $N$=10, as we have performed every control of VLC player and audio-player 10 times with the corresponding gesture commands.
\begin{figure*}[!ht]
\includegraphics[width=\textwidth,height=8 cm]{images/VLC_player.jpg}
\caption{\label{VLC_player} VLC player control using gesture commands.}
\end{figure*}

\subsection{\textbf{Audio player (Spotify) control:}}
\begin{figure*}[!ht]
\includegraphics[width=\textwidth,height=6.5 cm]{images/play_pause_audio.png}
\caption{\label{audio_player} Audio player (Spotify platform) control using gesture commands.}
\end{figure*}

A summary for controlling different functions of audio-player using various gesture commands has been tabulated in Table \ref{gesture_controls}, whereas Fig. \ref{audio_player} shows the demo of the gesture-controlled audio player. In our work, we have used the `audioplayer' \cite{pypi} python package and `playerctl' command-line library to use different features of the Spotify music player, such as playing a song, pausing, resuming, switching to the next song, and go to the previous music. We have executed every control of the Spotify player (such as: playing, pausing, resuming, etc.) ten times to obtain some valuable results. For instances shown in Table \ref{gesture_controls}, gesture `Ok' has been customized to play the song, similarly, `Five' for pause, `Fist' for resume, `Two' for going to the next song, and `Three' for a switch to the previous music. In Table \ref{VLC_audio_performance}, the performance of the gesture-controlled audio player (Spotify player) in real-time scenario, along with the number of hits, misses, and the average response time (in mili-seconds), has been depicted.

\begin{table}
\centering
\caption{\label{inference_speed}Real-time Performance comparison among all pre-trained CNN models and ViT in terms of inference speed (FPS), time (mili-sec), and model weight file size (.h5 format).}
\color{blue}
\label{inference_speed}
\resizebox*{0.50\textwidth}{!}{   
\begin{tabular}{|c|c|c|c|} 
\hline
Models       & \begin{tabular}[c]{@{}c@{}}Model weight\\file size (.h5 format)\\in MB\end{tabular} & \begin{tabular}[c]{@{}c@{}} FPS\end{tabular} & \begin{tabular}[c]{@{}c@{}}Inference \\~time (mili-sec)\end{tabular}  \\ 
\hline
VGG16        & 70                                                                                  & 18                                                              & 0.60                                                                  \\ 
\hline
VGG19        & 80                                                                                  & 16                                                              & 0.76                                                                  \\ 
\hline
Inception-V1 & 28                                                                                 & 25                                                              & 0.20                                                                  \\ 
\hline
ResNet-50    & \multicolumn{1}{l|}{~ ~ ~ ~ ~ ~  80}                                               & 18                                                              & 0.50                                                                  \\ 
\hline
ResNet-101   & \multicolumn{1}{l|}{~ ~ ~ ~ ~ ~  90}                                               & 15                                                              & 0.90                                                                  \\ \hline \hline
ViT          & 42                                                                                  & 20                                                              & 0.40                                                                  \\
\hline
\end{tabular}
}
\end{table}
\color{black}
\begin{table*}[!ht]
\caption{\label{gesture_controls}Desktop applications control by using different gesture commands.}
\begin{center}
    
\begin{tabular}{|cc|cc|cc|}
\hline
\multicolumn{2}{|c|}{\textbf{VLC player}}                       & \multicolumn{2}{c|}{\textbf{Audio player}}           & \multicolumn{2}{c|}{\textbf{2D Super-mario bros}}        \\ \hline
\multicolumn{1}{|c|}{Gesture label} & Action           & \multicolumn{1}{c|}{Gesture label} & Action & \multicolumn{1}{c|}{Gesture label} & Action     \\ \hline
\multicolumn{1}{|c|}{Ok}            & Play             & \multicolumn{1}{c|}{Ok}            & Play   & \multicolumn{1}{c|}{Ok}            & Run        \\ \hline
\multicolumn{1}{|c|}{L}             & Volume Up        & \multicolumn{1}{c|}{Five}          & Pause  & \multicolumn{1}{c|}{Five}          & Jump right \\ \hline
\multicolumn{1}{|c|}{Hang}          & Volume down      & \multicolumn{1}{c|}{Fist}           & Resume & \multicolumn{1}{c|}{Two}           & Hold/Stay  \\ \hline
\multicolumn{1}{|c|}{Close}         & Video quit       & \multicolumn{1}{c|}{Two}             & Go to next song   & \multicolumn{1}{c|}{Four}          & Jump left  \\ \hline
\multicolumn{1}{|c|}{Two}           & Go to next video &

\multicolumn{1}{c|}{Three}              &  Go to previous song & \multicolumn{1}{c|}{Thumb}                &   Jump           \\ \hline
\multicolumn{1}{|c|}{Five}          & Pause            & \multicolumn{1}{c|}{}              &        & \multicolumn{1}{c|}{}             &           \\ \hline
\end{tabular}
\end{center}
\end{table*}

\begin{table*}[!ht]
\caption{\label{VLC_audio_performance}Real-time performance analysis of  gesture-command-based VLC and Spotify player control (10 times).}
\begin{center}
\resizebox*{1\textwidth}{!}{   
\begin{tabular}{|cccccc|}
\hline
\multicolumn{6}{|c|}{Performance analysis of real-time gesture-command based VLC-player application (10 times)}                                                                                                                                  \\ \hline
\multicolumn{1}{|c|}{Action}           & \multicolumn{1}{c|}{Corresponding gestures} & \multicolumn{1}{c|}{Number of hits} & \multicolumn{1}{c|}{Number of  misses} & \multicolumn{1}{c|}{Detection rate (\%)} &  \begin{tabular}[c]{@{}c@{}}Average Response \\ time (ms)\end{tabular} \\ \hline
\multicolumn{1}{|c|}{Play}             & \multicolumn{1}{c|}{Ok}                     & \multicolumn{1}{c|}{10}             & \multicolumn{1}{c|}{0}                 & \multicolumn{1}{c|}{100}                 &      0.35                   \\ \hline
\multicolumn{1}{|c|}{Volume Up}        & \multicolumn{1}{c|}{L}                      & \multicolumn{1}{c|}{8}               & \multicolumn{1}{c|}{2}                  & \multicolumn{1}{c|}{80}                    &    0.83                     \\ \hline
\multicolumn{1}{|c|}{Volume down}      & \multicolumn{1}{c|}{Hang}                   & \multicolumn{1}{c|}{9}               & \multicolumn{1}{c|}{1}                  & \multicolumn{1}{c|}{90}                    &    0.82                  \\ \hline
\multicolumn{1}{|c|}{Quit}             & \multicolumn{1}{c|}{Close}                  & \multicolumn{1}{c|}{10}               & \multicolumn{1}{c|}{0}                  & \multicolumn{1}{c|}{100}                    &       0.38                   \\ \hline
\multicolumn{1}{|c|}{Go to next video} & \multicolumn{1}{c|}{Two}                    & \multicolumn{1}{c|}{7}               & \multicolumn{1}{c|}{3}                  & \multicolumn{1}{c|}{70}                    &      0.38                    \\ \hline
\multicolumn{1}{|c|}{Pause}            & \multicolumn{1}{c|}{Five}                   & \multicolumn{1}{c|}{9}               & \multicolumn{1}{c|}{1}                  & \multicolumn{1}{c|}{90}                    &   0.40                      \\ \hline \hline
\multicolumn{6}{|c|}{Performance analysis of Spotify player (10 times) based on gesture commands in real-time scenario}                                                                                                                     \\ \hline
\multicolumn{1}{|c|}{Play}             & \multicolumn{1}{c|}{Ok}                     & \multicolumn{1}{c|}{10}               & \multicolumn{1}{c|}{0}                  & \multicolumn{1}{c|}{100}                    &    0.30                      \\ \hline
\multicolumn{1}{|c|}{Pause}            & \multicolumn{1}{c|}{Five}                   & \multicolumn{1}{c|}{10}               & \multicolumn{1}{c|}{0}                  & \multicolumn{1}{c|}{100}                    &       0.30                   \\ \hline
\multicolumn{1}{|c|}{Resume}           & \multicolumn{1}{c|}{Fist}                    & \multicolumn{1}{c|}{9}               & \multicolumn{1}{c|}{1}                  & \multicolumn{1}{c|}{90}                    &     0.40                     \\ \hline
\multicolumn{1}{|c|}{Go to next song}             & \multicolumn{1}{c|}{Two}                      & \multicolumn{1}{c|}{8}               &     \multicolumn{1}{c|}{2}                  & \multicolumn{1}{c|}{80}                    &                0.35          \\ \hline
\multicolumn{1}{|c|}{Go to previous song}      & \multicolumn{1}{c|}{Three}                  & \multicolumn{1}{c|}{7}               & \multicolumn{1}{c|}{3}                  & \multicolumn{1}{c|}{70}                    &        0.35                  \\ \hline
\end{tabular}
}
\end{center}
\end{table*}

\subsection{\textbf{Super Mario-Bros game control}}

In this section, we have demonstrated the experimental details of the gesture-controlled super-Mario Bros game along with real-time performance analysis. Here we have used the OpenAI Gym environment and imported $`gym - super - mario-bros'$  \cite{gym-super-mario-bros} command-line interface to trigger different controls of this game based on various gesture commands. For example, the gesture `Ok' has been set for the run of Mario character; similarly, gestures `Four' and `Five' have been customized for jumping to the left and right direction, respectively, and gesture `Two' for hold/stay of the Mario. This game's controls are tabulated in Table \ref{gesture_controls}. One Demo of this game is shown in Fig. \ref{mario_bros}.
\begin{figure}[!ht]
\includegraphics[width=8 cm,height=7.5 cm]{images/mario_bros.jpg}
\caption{\label{mario_bros} 2D Mario-Bros game control using gesture commands.}
\end{figure}

\begin{figure}[!ht]
\includegraphics[width=7.5 cm,height=7.5 cm]{images/virtual_mouse2.png}
\caption{\label{virtual_mouse} Schematic diagram of virtual mouse.}
\end{figure}

\begin{figure}[!ht]
\centering
\includegraphics[width=6cm,height=16cm]{images/kalman_filter_tracking.jpg}
\caption{\label{kalman_filter} Block diagram for Kalman filter-based tracking.}
\end{figure} 

\begin{figure*}[!ht]
\includegraphics[width=\textwidth,height=7 cm]{images/virtual_mouse1.png}
\caption{\label{virtual_mouse1} Demo of gesture-controlled virtual mouse.}
\end{figure*}

\subsection{Building of virtual mouse}
In order to make the interface more convenient, we have developed a virtual mouse, where main objective is to control various functions of a traditional mouse with the help of predicted gesture commands instead of a physical mouse device. The main functions of a traditional mouse are click, right-click, double-click, drag, scroll wheel up/down and cursor movement. In our work, we have mimicked different events of a physical mouse with seven corresponding gesture commands. For instances, the gestures `Two' is set for right click, `Ok' is for double click, `Index' is for left click etc. Fig. \ref{virtual_mouse} shows the schematic diagram of the virtual mouse and one demo of gesture-controlled virtual mouse is depicted in Fig. \ref{virtual_mouse1}. The summary of different controls of virtual mouse according to customized gestures have been shown in Table \ref{mouse_control}. But still, there is a problem regarding the smooth movement of the mouse cursor, for example, during slight hand movement, jumping of cursor from one position to another on the screen. We have also resolved this issue by applying the Kalman filtering \cite{asaari2010hand}. In section \ref{mouse-cursor-kalman}, we have shown the performance analysis of improved mouse cursor control using Kalman filter.

\begin{table}[!ht]
\caption{\label{mouse_control}Triggering of mouse events using customized gesture inputs.}
\begin{center}
\begin{tabular}{|c|c|}
\hline
Gesture inputs & Action           \\ \hline
Two            & Right click      \\ \hline
Ok             & Double click     \\ \hline
Index          & Left click       \\ \hline
Five           & Pointer movement \\ \hline
Heavy          & Scroll up        \\ \hline
Hang           & Scroll down      \\ \hline
Palm           & Drag             \\ \hline
\end{tabular}

\end{center}

\end{table}

\subsubsection{\label{mouse-cursor-kalman} Kalman filter based mouse-pointer tracking} 
In tracking mouse-cursor movement, first, we need to get the coordinates of the detected hand portion, followed by calculating the centroid position. Centroid calculation is mandatory during hand movements because it is dynamic and changes with respect to time. \color{blue} Suppose the coordinates of the detected hand portion is (x,y,w,h), then the centroid will become (x+ $\frac{w}{2}$), (y+ $\frac{h}{2}$). After obtaining the coordinates, it will determine whether hand tracking is available or not; if it is, we immediately apply the Kalman filter; if not, we initialize the hand tracking. Next, it recursively estimates the next state and corrects the current state to minimize error, then uses the prediction as the mouse cursor's tracking path. We have selected the state space model for the process characterized by the movement of both x and y directions, which represent horizontal and vertical coordinates, respectively. Then we chose the pointer movement in a linear way and displayed the results based on the hand's movement from left to right and right to left when controlling the virtual mouse. The overall flow diagram of the Kalman filter-based virtual mouse pointer tracking is illustrated in Fig. \ref{kalman_filter}. 
We have found that employing the Kalman filter makes the gesture-controlled mouse pointer's movement very smooth, and it prevents the jumping of the cursor through the screen. \color{black} One demo of Kalman filter-based cursor movement is illustrated in Fig. \ref{virtual_mouse1},. One drawback of exploring the Kalman filter in our work is that it introduces some delay during a real-time scenario, which is acceptable.

\section{\label{discussion}Discussions}
An interactive HCI interface is suggested through a vision-based hand gesture classification system based on deep learning approach. We have performed several experiments on two datasets (details description in section \ref{dataset_desciption}) to evaluate the potency of our proposed work. We have computed the performance of five pre-trained CNN models and ViT approach for input size $(64 \times 64)$ and $(128 \times 128)$. The results have been illustrated in Table \ref{CNN1-compare1}. It is observed that the Inception-V1 model has outperformed the other CNN models and ViT model due to having an Inception module, where $(1 \times 1)$ convolution was used for the dimension reduction purpose. Simultaneously, using this architecture reduces the computation cost, and training time becomes faster. It is also seen in Table \ref{CNN1-compare1} that the time taken to train with the images of dimensions $(64 \times 64)$ pixel is less than the time taken to train the images of size $(128 \times 128)$. However, training with smaller images leads to a huge reduction in computational cost. So, during inferencing, the images are resized into the aspect ratio of $(64 \times 64)$. In Table \ref{CNN3-compare1}, we have shown that our proposed framework (gesture detection + segmentation + classification using Inception-V1) has outperformed the existing schemes presented by \cite{yingxin2016robust}, \cite{neethu2020efficient} in terms of accuracy (\%), precision (\%), recall (\%) and F1-score (\%) after evaluating on Dataset-B. The performance analysis of our all six used models (five pre-trained CNN models and ViT) in the real-time scenario is tabulated in Table \ref{inference_speed}. \color{blue} It is noticed in Table \ref{inference_speed} that the average inference speed by using the Inception-V1 model and ViT is higher than the other considered models, but Inception-V1 exhibits slightly better performance compared to ViT model, so during real-time inference task, we have considered the Inception-V1 model weight file (.h5 format) for designing a low-cost HMI to control three multimedia applications (VLC player, Spotify music player, and 2D-Super-Marios game) and one gesture-controlled virtual mouse in real-time scenarios. \color{black} Table \ref{VLC_audio_performance} depicts the performance analysis for each application in the real-time scenario. The advantages of our proposed system are as follows.

\begin{enumerate}

\item [a)] The average speed of this system is 25 fps (frames per second), which meets the perfect requirements for real-time app control.

\item[b)] This HCI interface is highly beneficial for interacting with desktop applications without touching any mouse or keyboard in a real-time scenario. So this system can be a preferred choice for older or physically disabled people.  
\end{enumerate}
Despite the advantages mentioned above of our proposed system, this system encounters some drawbacks during real-time gesture prediction in the cluttered background because it creates multiple contour regions leading to the segmentation error (failed to segment the gesture portion). Illumination variation and distance issues are also challenges in our proposed system. Another project issue is controlling the higher fps applications, as the average speed of our proposed scheme is 25 fps. Furthermore, using the Kalman filter delays the virtual mouse's performance a bit slow in a real-time scenario. We will explore some effective object detection techniques, such as Faster RCNN \cite{ren2015faster}, EfficientDet \cite{tan2020efficientdet}, etc., to detect the gestures in the messy background and will bring some modifications to the Kalman filtering algorithm to avoid the time delay appears during the real-time performance.
\color{blue}
\section{\label{ablation}Ablation Study}
We have performed the ablation study to check the effect of the model performances using different optimizers such as stochastic gradient descent (SGD), RMSProp and AdaGrad, etc., after evaluating on Dataset-B. It is observed 
in Table \ref{CNN3-compare2}, that training with Adam optimizer performs slightly better than the other used optimizers in terms of precision, recall, accuracy, and F-score values.
\color{black}
\begin{table*}
\centering
\caption{\label{CNN3-compare2} Ablation study results of used models using different optimizers on Dataset-B.}

\label{CNN4-compare4}
\begin{tabular}{|c|c|C|E|E|E|} 
\hline
Optimizer                                 & Used Methods / models                                                                                                                       & Testing Accuracy (\%)                  & Precision (\%) & Recall (\%) & F-score (\%)  \\ 
\hline

\multirow{6}{*}{\textbf{SGD}}               & VGG16                                                                                                & 99.23                                  & 99.25          & 99.20       & 99.23         \\ 
\cline{2-6}
                                                 &  VGG19                                                                                                & 99.10                                  & 99.20          & 99.10       & 99.15         \\ 

\cline{2-6}
                                                 &  ResNet50                                                                                             & 98.42                                  & 99.14          & 99.20       & 99.17         \\ 
\cline{2-6}
                                                 &  ResNet101                                                                                            & 98.60                                  & 98.80          & 98.30       & 98.55         \\

                                                 \cline{2-6}
                                                 & Inception-V1                                                                                         & 99.25                                  & 99.28          & 99.22       & 99.25         \\ 
\cline{2-6}
                                                 &  ViT                                                                                                  & 93.57                                  & 94.60          & 94.80       & 94.70         \\ 
\hline 
\hline
\multirow{6}{*}{\textbf{RMSprop}}               & VGG16                                                                                                & 98.51                                  & 98.56          & 98.28       & 98.42         \\ 
\cline{2-6}
                                                 &  VGG19                                                                                                & 98.75                                  & 98.94          & 99.00       & 98.97         \\ 

\cline{2-6}
                                                 &  ResNet50                                                                                             & 96.85                                  & 97.10          & 97.25       & 97.18         \\ 
\cline{2-6}
                                                 &  ResNet101                                                                                            & 98.15                                  & 98.30          & 98.20       & 98.25         \\

                                                 \cline{2-6}
                                                 & Inception-V1                                                                                         & 98.95                                  & 99.00          & 98.80       & 98.90         \\ 
\cline{2-6}
                                                 &  ViT                                                                                                  & 99.05                                  & 98.90          & 98.80       & 98.85         \\
\hline
\hline
\multirow{6}{*}{\textbf{AdaGrad}}               & VGG16                                                                                                & 95.20                                  & 95.25          & 95.60       & 95.43         \\ 
\cline{2-6}
                                                 &  VGG19                                                                                                & 98.55                                  & 98.10          & 98.30       & 98.20         \\ 

\cline{2-6}
                                                 &  ResNet50                                                                                             & 92.45                                  & 92.30          &       92.50 & 92.40         \\ 
\cline{2-6}
                                                 &  ResNet101                                                                                            & 92.15                                  & 92.30          & 92.20       & 92.25         \\

                                                 \cline{2-6}
                                                 & Inception-V1                                                                                         & 96.25                                  & 96.50          & 96.10       & 96.30         \\ 
\cline{2-6}
                                                 &  ViT                                                                                                  & 94.50                                  & 94.40          & 94.30       & 94.35         \\
\hline

\end{tabular}
\end{table*}

\section{\label {conclusion} Conclusion}
 In this paper, a real-time HCI to control two desktop applications (VLC player and Spotify music player), and one 2D game, along with the development of the virtual mouse using gesture commands, has been depicted. One public and one custom datasets have been considered to validate our proposed approach. We have compared the performance of five pre-trained CNN models and ViT on Dataset-A and Dataset-B. The results show that the Inception-V1 model has achieved better results in terms of training time and validation accuracy (\%). The comparative analysis with other proposed schemes also shows that the Inception-V1 model exhibited superior results and was chosen for the real-time gesture classification task for faster training time (sec) and also better accuracy (\%). We have also discussed the real-time performance analysis of every gesture-controlled application and depicted the average response time (ms) for every control of every application. Moreover, we have applied the Kalman filter to make the motion of the gesture-controlled mouse cursor smoother. In our future work, we will build a robust and efficient HCI interface by connecting other modalities, such as eye-gaze tracking, facial expression, etc., to make the system more convenient and user-friendly.


%
%


\newpage





\bibliographystyle{elsarticle-num}
\bibliography{ref}

\appendix
\section{Appendix}\label{appendix-1}
\subsection{\textbf{VGG16}}
 VGG16 architecture is an one type of variant of VGGNet \cite{simonyan2014very} model, consisting of 13 convolutional layers with kernel sizes of $(3 \times 3)$ and Relu activation function. Each convolution layer is pursued by max-pooling layer with filter size $(2 \times 2)$. Finally the FC layer is added with softmax activation function to produce the final output class label. In this architecture, the depth of network is increased by adding more convolution and max-pooling layers. This network is trained on large-scale ImageNet \cite{deng2009imagenet} dataset, ImageNet dataset  consists of millions of images and having more than 20,000 class labels developed for large scale visual recognition challenge. VGG16 has reported test accuracy of 92.7\% in the ILSVRC-2012 challenge.

\subsection{\textbf{VGG19}}
VGG19 is also a variant of VGGNet network \cite{simonyan2014very}, it comprises 16 convolutional layers and three dense layers with kernel /filter size of $(3 
\times 3)$ and then the max-pooling layer is used with filter size $ (2 \times 2)$ . This architecture is pursued by the final FC layer with softmax function to deliver the predicted class label. This model has achieved second rank in ILSVRC-2014 challenge after being trained on ImageNet \cite{deng2009imagenet} dataset. This model has an input size of $(224 \times 224)$.

\subsection{\textbf{Inception-V1}}
Inception-V1 or GoogleNet \cite{szegedy2015going} is a powerful CNN architecture with having 22 layers built on the inception module. In this module, the architecture is limited by three independent filters such as $(1 \times 1)$, $(3 \times 3)$ and $(5 \times 5)$. Here in this architecture, $(1 \times 1)$ filter is used before 
$(3 \times 3)$, $(5 \times 5)$ convolutional filters for dimension reduction purpose. This module also includes one max-pooling layer with pool size $(3 \times 3)$. In this module the outputs by using convolutional layers such as $(1 \times 1)$, $(3 \times 3)$ and $(5 \times 5)$ are concatenated and form the inputs for next layer. The last part follows the FC layer with a softmax function to produce the final predicted output. This input of this model is $(224 \times 224)$. This architecture is trained with ImageNet \cite{deng2009imagenet} dataset and has reported top-5 error of 6.67\% in ILSVRC-2014 challenge.

\subsection{\textbf{ResNet50:}} 
Residual neural network \cite{he2016deep} was developed by Microsoft research , This model consists of 50 layers, where 50 stands total number of deep layers, containing 48 convolutional layers, one max-pooling. Finally global average pool layer is connected to the top of the final residual block, which is pursued by the dense layer with softmax activation to generate the final output class. This network has input size of $(224 \times 224)$. The backbone of this architecture is based on residual block. In the case of residual block, the output of one layer is added to a deeper layer in the block, which is also called skip connections or shortcuts. This architecture also reduces the vanishing and exploding gradient problems during training. ResNet50 architecture was trained on the ImageNet dataset \cite{deng2009imagenet} and has achieved a good results in ILSVRC-2014 challenge with an error of 3.57\%.

\subsection{\textbf{ResNet101:}}
 
 ResNet101 model consists of 101 deep layers. Like ResNet50, this architecture is also based on the residual building block. In our experiment, we have loaded the pre-trained version of this architecture, trained on ImageNet dataset \cite{deng2009imagenet} that comprises millions of images. This model's default input image size is $(224 \times 224)$.

\subsection{\textbf{Vision Transformer:}}
 
A standard transformer architecture consists of two components (1) a set of encoder and (2) a decoder. But in the case of ViT , it doesn't require the decoder part as it contains only encoder part. In Vision Transformer, firstly image is split into fixed-sized patches, and each patch is implemented for the patch-embedding phase. In the case of patch embedding, each patch is flattened to produce a one-dimensional vector. After the patch-embedding phase, positional embedding is added with the patches to retain the positional information about the image patches in the sequence. Next, they are moved to the transformer encoder. The transformer encoder \cite{bazi2021vision} comprises two components: (1) a Multi-head self-attention block (MHSA) and (2) MLP (multiple-layer perceptron). Hence MHSA block splits the inputs into several number of heads so that each head can learn different levels of self-attention. Then, the outputs of multiple attention heads are concatenated and delivered to the MLP. Next, the classification task is performed by the MLP layer.

\section{Appendix}\label{appendix-2}
\subsection{\label{statistical} Statistical hypothesis testing}
We have also performed a statistical analysis in order to check statistical significance of our model. In Experiment-1 \ref{experiment1}, we have conducted one sample $t$-test with the help of IBM SPSS statistical analysis tool. 
In case of null hypothesis, we have to assume that our model is not statistically significant. \\
To obtain the value of $t$, the following formula is used: 
\begin{center}
t=$\frac{(\overline{X}-\mu)}{\frac{SD}{\sqrt{k}} } $    \end{center}
\begin{table}[!ht]
\caption{\label{one-sample} One sample statistics.}
\centering
\resizebox*{0.5\textwidth}{!}{   
\begin{tabular}{cccc}
\\ \hline
\begin{tabular}[c]{@{}c@{}}Number of \\  samples\end{tabular} & Mean    & Standard deviation & Standard error mean \\ \hline
10                & 99.8300 & 0.2907             & 0.0919  \\ \hline
\end{tabular}
}
\end{table}
\begin{table*}[!ht]
\caption{\label{one-sample-test} One-sample T-test result.}
\centering
\resizebox*{0.75\textwidth}{!}{   
\begin{tabular}{ccccccc}
\hline
\multicolumn{7}{c}{\textbf{One-sample Test}}                    \\\hline
      &    & \multicolumn{2}{c}{Significance}   & Test value= 99  & \multicolumn{2}{c}{\multirow{2}{*}{\begin{tabular}[c]{@{}c@{}}95\% Confidence interval \\ of the difference\end{tabular}}} \\
      &    & \multicolumn{2}{c}{}               &                 & \multicolumn{2}{c}{}                                                                                                       \\ \hline
t     & df & One-Sided p     & Two-Sided p      & Mean Difference & Lower                                                        & Upper                                                       \\
9.026 & 9  & \textless 0.001 & \textless{}0.001 & 0.8300           & 0.6219                                                       & 1.038  \\ \hline                                                    
\end{tabular}
}
\end{table*}
Where $\overline{X}$ is the mean of samples. \\
$\mu$\ the test value. \\
SD sample standard deviation. \\
k size of samples. \\
To get the value of $\overline{X}$, firstly, we have used the ten-fold-cross-validation strategy using Dataset-1, and have calculated the fold-wise accuracy followed by computing the average (considered as sample mean) of these accuracy values with the Inception-V1 (best-selected model for Experiment-1 \ref{experiment1}) model.\\
Table \ref{one-sample} shows that the sample mean ($\overline{X}$), sample size (k), test value ($\mu$), and the standard deviation (SD) are 99.83, 10, 99, and 0.2907, respectively, and the entire statistical analysis of one-sample t-test has been demonstrated in Table \ref{one-sample-test}.\\
The results in Table \ref{one-sample-test} exhibit that $p$-value  \textless 0.001. Here $p$-value is used for hypothesis testing to determine whether there is evidence to reject the null hypothesis. \\
If $p$ \textless $\alpha$ where, $\alpha$ (confidence level) = 0.05, then null hypothesis is rejected. \\ In Table \ref{one-sample-test},
it is observed that $p$-value is very less than $\alpha$, so the null hypothesis is rejected, and
we can say that there is a statistically significant difference in the mean of the accuracy values.